\pdfoutput=1

\documentclass[11pt]{article}

\usepackage[]{acl}

\usepackage{times}
\usepackage{latexsym}

\usepackage{times}
\usepackage{latexsym}
\usepackage{hyperref}
\usepackage{url}
\usepackage{times}
\usepackage{latexsym}
\usepackage{graphicx}
\usepackage{enumitem}
\usepackage{subfigure}
\usepackage{makecell}
\usepackage{amssymb}
\usepackage{lipsum}
\usepackage{booktabs}
\usepackage[normalem]{ulem}
 \usepackage{mathrsfs}
\usepackage{multirow}
\usepackage{colortbl}
\usepackage{tabularx}
\usepackage{amsmath}
\usepackage{framed}
\usepackage{bbding}
\usepackage{lipsum}
\usepackage{longtable}
\usepackage{diagbox}
\usepackage{fancyhdr}
\usepackage{multicol}
\usepackage{algorithm}
\usepackage{algorithmic}
\usepackage[T1]{fontenc}

\usepackage[utf8]{inputenc}

\usepackage{microtype}

\usepackage{inconsolata}

\title{Meta-Reasoning: Semantics-Symbol Deconstruction\\ for Large Language Models}

\author{Yiming Wang$^1$ ~~Zhuosheng Zhang$^1$ ~~Pei Zhang$^{2,}$\thanks{~~~Rui Wang and Pei Zhang are Co-corresponding Authors.} ~~Baosong Yang$^2$ ~~Rui Wang$^{1,*}$ \\
        $^1$Shanghai Jiao Tong University \quad
        $^2$Alibaba Group Inc. \\
        \texttt{\{yiming.wang,zhangzs,wangrui12\}@sjtu.edu.cn} \\
        \texttt{psyangqi@gmail.com}, \quad
        \texttt{yangbaosong.ybs@alibaba-inc.com}}

\begin{document}
\maketitle
\begin{abstract}
Neural-symbolic methods have demonstrated efficiency in enhancing the reasoning abilities of large language models (LLMs). However, existing methods mainly rely on syntactically mapping natural languages to complete formal languages like Python and SQL. 
Those methods require that reasoning tasks be convertible into programs, which cater to the computer execution mindset and deviate from human reasoning habits.
To broaden symbolic methods' applicability and adaptability in the real world, we propose the \textbf{Meta-Reasoning} from a linguistic perspective. This method empowers LLMs to deconstruct reasoning-independent semantic information into generic symbolic representations, thereby efficiently capturing more generalized reasoning knowledge.
We conduct extensive experiments on more than ten datasets encompassing conventional reasoning tasks like arithmetic, symbolic, and logical reasoning, and the more complex interactive reasoning tasks like theory-of-mind reasoning.
Experimental results demonstrate that Meta-Reasoning significantly enhances in-context reasoning accuracy, learning efficiency, out-of-domain generalization, and output stability compared to the Chain-of-Thought technique.
Code and data are publicly available at \url{https://github.com/Alsace08/Meta-Reasoning}.
\end{abstract}

\section{Introduction}

Symbols serve as the primitive carrier through which humans can comprehend, articulate, and conceptualize the intricacies of both nature and society \citep{peirce1902logic}.
From a cross-linguistic perspective, ideographic symbolic languages like Arabic numerals, mathematical symbols, and emojis can transcend barriers to natural semantic understanding. They serve as a universal representation across ethnically diverse human languages \citep{chen2022program,cheng2022binding,wei2023symbol,liu2023zero,das2023evaluating}, facilitating communication and comprehension on a global scale.
In a specific mono-linguistic communication scenario, symbols inherently possess multiple referential meanings shaped by social and cultural properties \citep{blumer1986symbolic}.
Consequently, a single symbol can encapsulate diverse semantic representations.
Conversely, various semantic representations can converge onto the same symbol, forming a many-to-one relationship when abstracting referential meanings.
This transformation opens avenues for transforming natural language reasoning into more generalized patterns, enabling efficient solutions.

\begin{figure}
  \centering
  \includegraphics[width=0.5\textwidth]{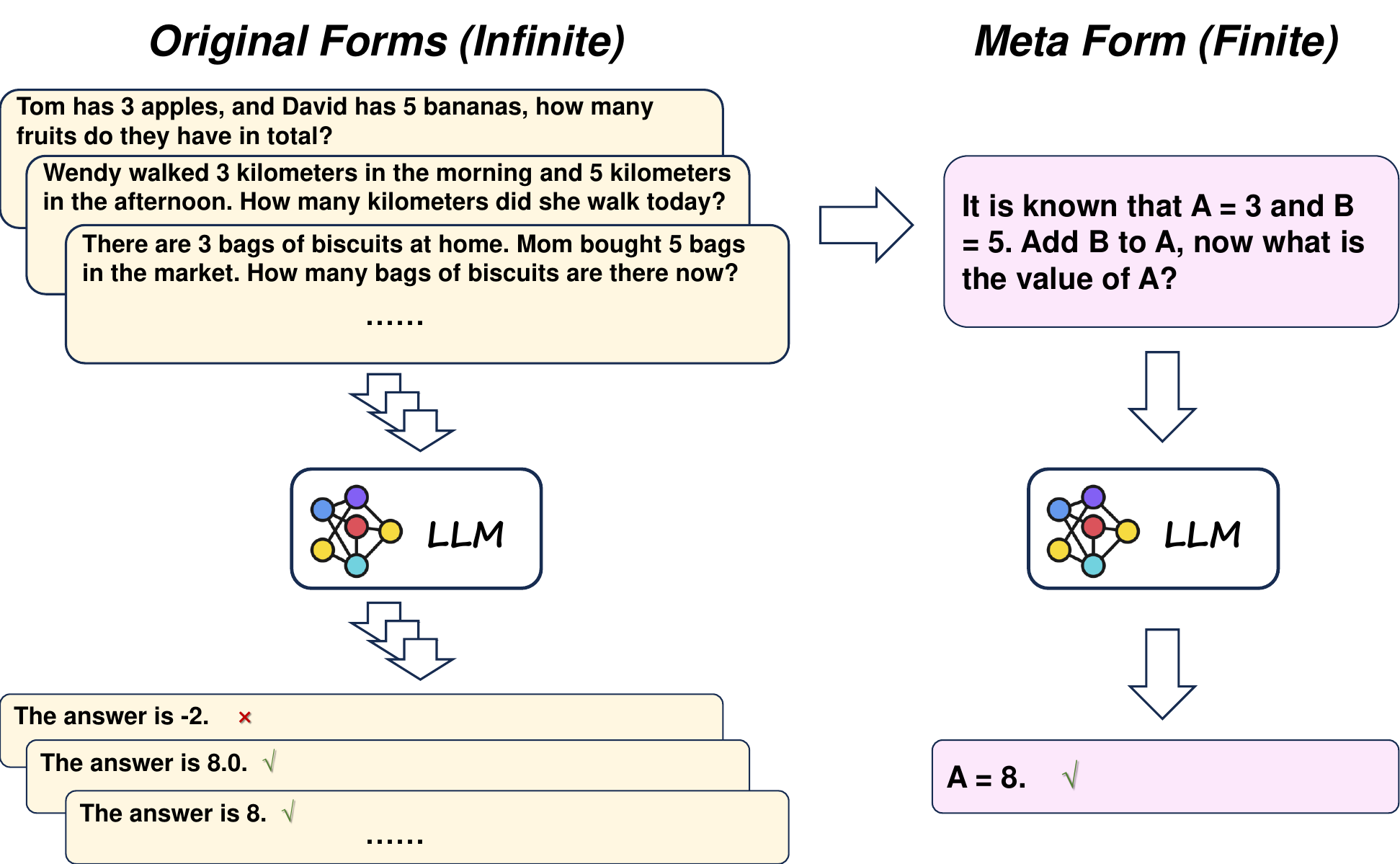}  
  \vspace{-3mm}
  \caption{Numerous language reasoning tasks exhibit meta-forms, wherein identifying general patterns can alleviate the reasoning burden on LLMs and facilitate learning through analogy.}
  \label{img:mr_total}
\vspace{-3mm}
\end{figure}

Current reasoning paradigms of large language models (LLMs), such as Chain-of-Thought (CoT) prompting \citep{weichain,kojima2022large,zhang2022automatic}, rely on multiple in-context learning demonstrations to perform well. However, the number of demonstrations is limited by LLMs' input capacity and inference cost, rendering it impractical to cover the distribution of specific task features exhaustively.
Therefore, we advocate a paradigm shift from infinite semantics systems to finite symbolic systems so that LLMs can acquire more generic knowledge with enhanced data learning efficiency, as shown in Figure \ref{img:mr_total}.

Motivated by the insight above, we introduce \textbf{Meta-Reasoning}, a novel reasoning paradigm aimed at deconstructing the semantics of entities and operations in questions into generic symbolic representations.
Meta-Reasoning enables LLMs to learn generalized reasoning patterns across various semantics-wrapped scenarios, enhancing learning efficiency and reasoning accuracy.
We apply Meta-Reasoning to in-context learning by designing demonstrations integrating semantic resolution with the CoT technique. This empowers LLMs to deconstruct questions and effectively capture more generalized knowledge autonomously.

To assess the efficacy of our method, we conduct experiments on over ten datasets, spanning both conventional reasoning scenarios, which involve arithmetic, symbolic, and logical reasoning tasks, and interactive reasoning scenarios, which involve theory-of-mind reasoning.
We mainly compare our method with the CoT method upon GPT-3 and ChatGPT.
Experimental results show that Meta-Reasoning consistently outperforms the Few-Shot-CoT method across all tasks, demonstrating significant performance improvements.
In the conventional reasoning scenarios, Meta-Reasoning achieves an average performance gain of +20\% across all datasets with fewer demonstrations.
In more complex interactive reasoning scenarios, Meta-Reasoning surpasses CoT across all levels of theory-of-mind reasoning with just a single demonstration.
Moreover, Meta-Reasoning demonstrates remarkable out-of-domain generalization and output stability, indicating its scalability and user-friendly nature as a reasoning paradigm.

To our knowledge, we are the first to establish an equivalence mapping from semantics to symbols within natural language. This innovation facilitates in-context learning for LLMs, significantly enhancing their capacity for generalized reasoning.
We expect to extend the reasoning ability boundary of LLMs based on this research.
\section{Preliminary: Why Meta-Reasoning?}

Meta-Reasoning is an idealized reduction-based reasoning paradigm defined in this work, whose goal is to reduce the infinite semantic concepts in the world's languages to a finite symbolic system, thus allowing machines to generalize to many semantically wrapped problems through the acquisition of universal laws.
This paradigm is best suited for such a reasoning scenario: the final reasoning results are independent of the particular semantic representations and are only related to the underlying reasoning skeletons.

The core of Meta-Reasoning lies in \textbf{Semantic-Symbolic Deconstruction}, which we simplify as \textbf{Semantic Resolution}.
This process conveys the semantics of the original problem via symbols with generalized meanings, without affecting the final results.
However, \textit{why deploying Semantic Resolution in LLMs} is a key issue, we must consider the advantages it brings to the reasoning process.

We explore this issue from two perspectives: (i) the human reasoning speed when responding to different questions, and (ii) the machine reasoning accuracy when responding to different questions.
We select MultiArith \citep{roy2015solving} and GSM8K \citep{cobbe2021training}, two arithmetic datasets, and rephrase 100 questions in each dataset according to the semantic resolution rules that will be introduced in Section \ref{subsec:deconstruction}, thereby creating meta-questions.
Subsequently, we distribute the original and meta-questions to both human volunteers and LLMs to obtain corresponding results of metrics.

\subsection{Response Speed Test For Human}

\begin{figure}[htb]
  \centering
  \includegraphics[width=0.96\columnwidth]{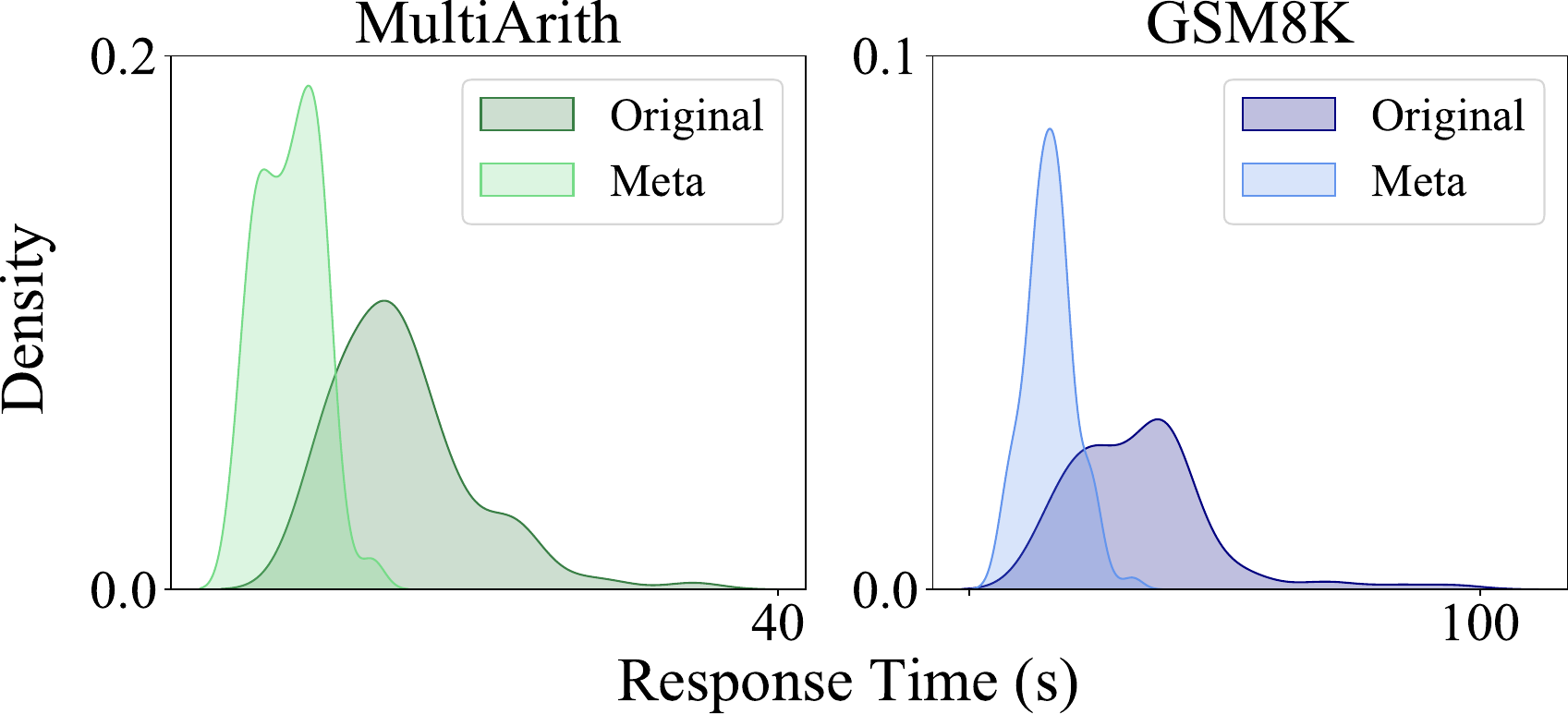}  
  \vspace{-0.05in}
  \caption{Human response time comparisons when solving original and meta-questions.}
  \label{img:response}
\end{figure}

\begin{figure*}
  \centering
  \includegraphics[width=0.98\textwidth]{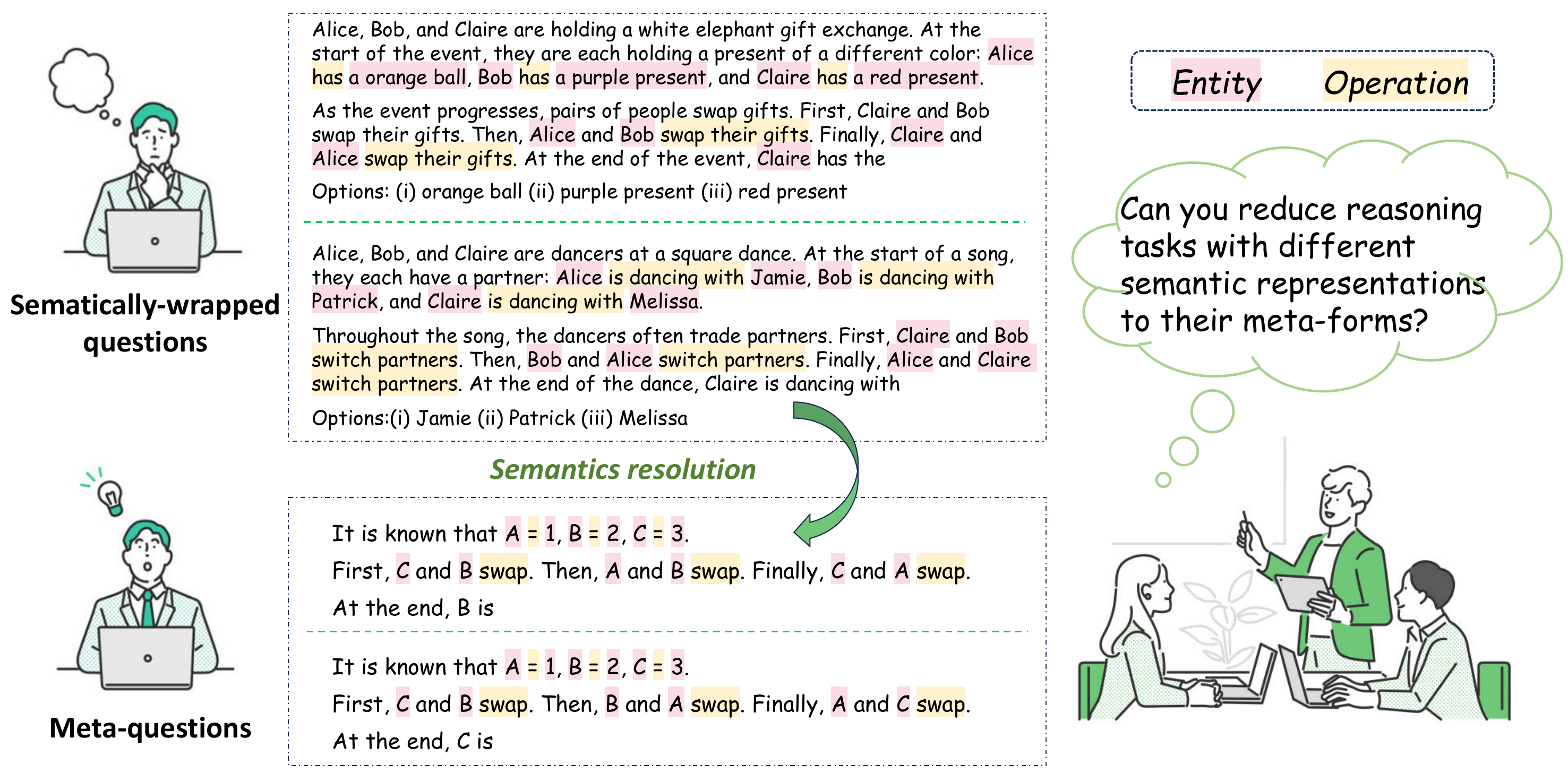}  
  \vspace{-0.1in}
  \caption{\textbf{Semantic Resolution of Meta-Reasoning}. We set resolution rules for \colorbox{pink}{Entity} and \colorbox{yellow}{Operation}.}
  \label{img:sr}
\end{figure*}

We assess the response speed of three human volunteers by measuring the total time taken from receiving the question to providing the answer.\footnote{Samples with incorrect answers are excluded from the analysis due to their negligible impact, given the low difficulty level of the math problems for adults.}
As shown in Figure \ref{img:response}, human response speed significantly improves when solving meta-questions, particularly evident in GSM8K.
This acceleration is attributed to the removal of unimportant semantic information in \textbf{meta-questions, which enables quicker recognition of the reasoning skeleton by humans}.
Moreover, the more concentrated distribution of human reaction times suggests a similarity in reasoning frameworks for such problems, indicating that \textbf{semantic resolution fosters consistency in reasoning patterns}.

\subsection{Accuracy Test For Machine}

\begin{table}[htb]
\begin{center}
\small
\resizebox{0.82\columnwidth}{!}
{
\begin{tabular}{lr}

\toprule
\multicolumn{2}{c}{\textit{MultiArith} (original $\rightarrow$ meta)} \\
\midrule
Zero-Shot & 28\% $\rightarrow$ 31\% (\textcolor{green}{$\mathrm{+}$}) \\
Zero-Shot-CoT & 70\% $\rightarrow$ 100\% (\textcolor{green}{+}) \\

\midrule
\multicolumn{2}{c}{\textit{GSM8K} (original $\rightarrow$ meta)} \\
\midrule
Zero-Shot & 22\% $\rightarrow$ 13\% (\textcolor{red}{-}) \\
Zero-Shot-CoT & 41\% $\rightarrow$ 97\% (\textcolor{green}{+}) \\

\bottomrule

\end{tabular}
}
\vspace{-0.0in}
\caption{LLMs Performance comparisons when solving original and meta-questions.}
\label{tab:pre}%
\end{center}
\end{table}

We assess the reasoning accuracy of GPT-3 using two prompting paradigms: standard Zero-Shot\footnote{The prompt is ``\textit{A:}''.} and Zero-Shot-CoT\footnote{The prompt is ``\textit{A: Let's think step by step.}''.} \citep{kojima2022large}.
As shown in Table \ref{tab:pre}, The Standard Zero-Shot method performs similarly on both types of questions, with notably poor performance on the GSM8K dataset. 
However, Zero-Shot-CoT yields markedly different outcomes. Specifically, when applied to the meta-questions, Zero-Shot-CoT demonstrates a significant performance improvement, particularly evident in the GSM8K dataset.
This observation suggests that \textbf{CoT reasoning for LLMs becomes notably smoother when tackling meta-problems}.

\section{Meta-Reasoning Paradigm}
\label{sec:mr}

We have observed notable performance improvements in LLMs when tackling questions after semantic resolution in the last section. 
In this section, we formally introduce the Meta-Reasoning paradigm employed in LLMs.
Section \ref{subsec:deconstruction} defines the specific rules for semantic resolution.
Then, we put this process through in-context learning for LLMs to imitate, and Section \ref{subsec:incontext} formalizes the demonstration design form of in-context learning.

\subsection{Definition: Semantic Resolution Rules}
\label{subsec:deconstruction}

Semantic resolution corresponds to the many-to-one mapping from various semantic representations to the most intrinsic symbolic representation.
We focus on two types of elements within text sequences that structure the entire reasoning skeleton but whose semantics do not change the reasoning path:
(i) \textbf{Entity}, it represents the subjects on which the reasoning task acts, but it is not critical what or who exactly it is;
(ii) \textbf{Operation}, it establishes connections and changes between subjects, but the exact form of that is not important. For example, ``he ate 3 apples'' and ``he threw 3 apples'' are both essentially forms of subtraction.
Examples are shown in Figure \ref{img:sr}.

\begin{figure*}
  \centering
  \includegraphics[width=\textwidth]{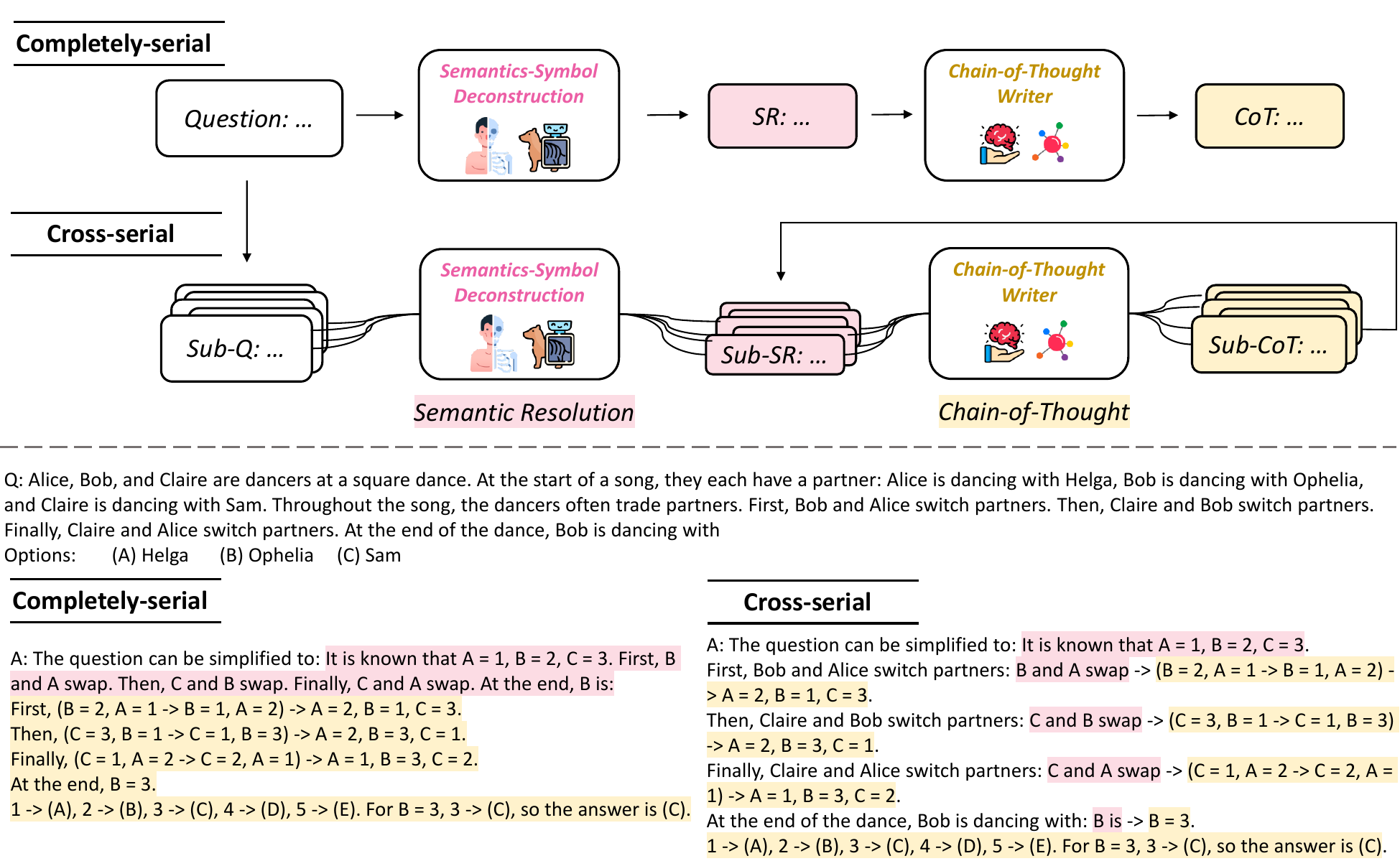}  
  \vspace{-0.18in}
  \caption{\textbf{In-context Learning Pipeline (Upper) and Example (Lower) of Meta-Reasoning}. The examples are taken from the Tracking Shuffled Objects task. For drafted demonstrations, we propose completely-serial and cross-serial fusion modes of semantic resolution and chain-of-thought, allowing LLMs to perform single-step reasoning more data-efficiently.}
  \label{img:pipeline}
  \vspace{-0.07in}
\end{figure*}

\paragraph{Entity.}
Intuitively, entity representations with natural semantics can be treated as the expansion products of an exhaustive set of non-empty symbols. 
Given a native symbol set\footnote{Take examples in the English language system, regardless of lowercase or uppercase.}(alphabet), ${\displaystyle \Sigma}^1 = \{A, B,..., Z\}$, the positive closure ${\displaystyle \Sigma}^+ = \bigcup_{i=1}^{\infty} {\displaystyle \Sigma}^i$ of ${\displaystyle \Sigma}^1$ contains the set $Q$  of all symbolic representations with natural semantics in the English language system, i.e., $Q \subset {\displaystyle \Sigma}^+$, where ${\displaystyle \Sigma}^i(i>1) = {\displaystyle \Sigma}^j \times {\displaystyle \Sigma}^{i-j} (1 \leq j \leq i)$ and $\times$ denotes the Cartesian product operation.
We consider the opposite form of the symbol-semantics expansion, i.e., semantics-symbol resolution, and construct the mapping $f_e: Q \rightarrow {\displaystyle \Sigma}^1$ to transform these complex semantic representations to their primitive symbolic form in the alphabet.
Since the symbols in the alphabet are meaningless, the mapping results are not required to be specified --- we default to mapping them one by one in alphabetical order without duplication.\footnote{For example, there are three semantic representations $x_1, x_2, x_3$ that need to be mapped, and the mapping can be done by default as $x_1 \rightarrow A, x_2 \rightarrow B, x_3 \rightarrow C$.}

Back to reasoning scenarios, given a sequence of original question $S = [s_{1:n}]$, we first manually locate all the entity spans $[s_{i:j}] \subset S$ (e.g. apple, mom), and later apply the mapping $f_e$ to them to obtain the single characters $\sigma_{ij} = f_e([s_{i:j}])$, respectively, which will be embedded back into the original position of the sequence $S$ so that it will be modified into $S = [s_{1:i-1} \circ \sigma_{ij} \circ s_{j:n}]$.

\paragraph{Operation.} 
Entities constitute the set of subjects on which the reasoning task acts, while the definition and change of entity states determine the reasoning path:
(i) \textit{definitions} of entity states can usually be reduced to assignment and logical association operations, \emph{i.e.}, $O_1 = \{=, \rightarrow\}$, and $O_1$ is a finite set;
(ii) \textit{changes} in entity states can be reduced to arithmetic operations, \emph{i.e.}, $O_2 = \{+,-,\times,\div\}$, and $O_2$ is a finite set.\footnote{There may be some extraordinary operations, but generally finite. We leave this for future work.}
Conveniently, these arithmetic symbols can correspond to natural semantics, e.g., ``+'' corresponds to ``add'', which allows symbols to be more closely integrated with natural language. 
Similar to the resolution of entities, we construct the mapping $f_o: Q \rightarrow (O_1 \cup O_2)$, and transform all manually-located operation representation $[s_{i:j}]$ (e.g. eat, have) into single symbols $\rho_{ij} = f_o([s_{i:j}])$, which will be embedded into the original position of the sequence $S$ so that it will be modified into $S = [s_{1:i-1} \circ \rho_{ij} \circ s_{j:n}]$.

Appendix \ref{sec:rulebase} provides some mapping examples.
After semantic resolution, the original questions maximally remove semantically irrelevant terms and simplify the need for semantic reasoning.

\subsection{Deployment: Synthetic Demonstration Design for In-context Learning}
\label{subsec:incontext}

However, manual annotation of entities and operations one by one is time-consuming.
We expect LLMs to autonomously learn generic reasoning patterns for certain reasoning tasks by automatically simplifying complex questions into equivalent and simpler forms. This can drive data-efficient learning.
Therefore, we consider the in-context learning.
Furthermore, inspired by the demonstrated significance of the CoT technique in enhancing reasoning capabilities in prior works \citep{wei2023symbol,kojima2022large}, we are dedicated to devising a fusion strategy of semantic resolution and CoT, which aims to maximize the performance potential of LLMs in reasoning.

We focus on two fusion modes: \textbf{Completely-serial} and \textbf{Crossly-serial}.
The primary distinction between the two modes lies in whether Semantic Resolution (SR) and CoT appear overlappingly.
The pipeline and case are illustrated in Figure \ref{img:pipeline}, with further details provided below:

\paragraph{Completely-serial.}
We first conduct SR to obtain the meta-question form, then draft the CoT for the corresponding meta-question.
In this case, the rationale is $\mathsf{[SR \circ CoT]}$.

\paragraph{Crossly-serial.}
We first split the original question into $n$ sub-steps, where $n$ may vary depending on the specific context.
For each sub-step $i$, the sub-rationale is represented as $\mathsf{[SR_i \circ CoT_i]}$.
Finally, we concatenate all the sub-rationales. 
In this case, the rationale is $\mathsf{[[SR_1 \circ CoT_1] \circ [SR_2 \circ CoT_2] \circ \cdots \circ [SR_n \circ CoT_n]]}$,
where $\mathsf{[SR_1 \circ SR_2 \circ \cdots \circ SR_n] = SR}$ and $\mathsf{[CoT_1 \circ CoT_2 \circ \cdots \circ CoT_n] = CoT}$.

\section{Experiments}
\label{sec:experiment}

\begin{table*}[tb]
\begin{center}
\setlength{\tabcolsep}{2.36mm}\small
{
\begin{tabular}{lcccccccc}

\toprule
\multirow{2}{*}{\bf Method} & \multicolumn{2}{c}{\bf Arithmetic} & \multicolumn{2}{c}{\bf Symbolic} & \multicolumn{3}{c}{\bf Logical} & \multirow{2}{*}{\bf Avg.}\\
\cmidrule(r){2-3}
\cmidrule(r){4-5}
\cmidrule(r){6-8}
& MultiArith & AddSub & Letter & Coin & Lies & Track(3/5/7) & Track(\textit{Avg.}) \\

\midrule

 \multicolumn{9}{c}{\bf Previous Fine-tuned SOTA} \\

\midrule
\rowcolor{gray!20}
 \multicolumn{9}{l}{\textit{Fine-tuned Paradigm}} \\
State-of-the-Art & 60.5 & 84.0 & - & - & 59.6 & - & 24.1 & - \\
\midrule

\multicolumn{9}{c}{\bf 175B GPT-3 (\texttt{text-davinci-002})} \\

\midrule
\rowcolor{gray!20}
 \multicolumn{9}{l}{\textit{Standard Prompting Paradigm}} \\
 Zero-Shot & 22.7 & 77.0 & 0.2 & 53.8 & 47.2 & 24.4 / 15.2 / 7.6 & 15.7 & 31.0\\
 Few-Shot & 33.8 & \underline{83.3} & 0.2 & 57.2 & 51.6 & -  & 25.1 & 37.7\\
 
 \rowcolor{gray!20}
 \multicolumn{9}{l}{\textit{Chain-of-Thought Paradigm}} \\
 Zero-Shot & 78.7 & 74.7 & 57.6 & 91.4 & 58.4 & 44.8 / 35.6 / 26.0 & 35.5 & 58.4\\
 Few-Shot & \underline{91.7} & 81.3 & \underline{59.0} & \underline{97.2} & \underline{92.0} & \underline{62.8} / \underline{60.8} / \underline{59.6} & \underline{61.1} & \underline{75.6} \\

  \rowcolor{gray!20}
 \multicolumn{9}{l}{\textit{Meta-Reasoning Paradigm (Ours)}} \\
 Few-Shot & {\bf 94.5} & {\bf 86.6} & {\bf 86.0} & {\bf 100.0} & {\bf 99.2} & {\bf 97.2 / 100.0 / 99.2} & {\bf 98.8} & {\bf 95.3}\\

 \midrule
 $\Delta$ & \textcolor{blue}{+2.8} & \textcolor{blue}{+3.3} & \textcolor{blue}{+27.0} & \textcolor{blue}{+2.8} & \textcolor{blue}{+7.2} & \textcolor{blue}{+34.4} / \textcolor{blue}{+39.2} / \textcolor{blue}{+39.6} & \textcolor{blue}{+37.7} & \textcolor{blue}{+19.7}\\
 \midrule

\multicolumn{9}{c}{\bf 175B GPT-3 (\texttt{text-davinci-003})} \\

\midrule
 
 \rowcolor{gray!20}
 \multicolumn{9}{l}{\textit{Chain-of-Thought Paradigm}} \\
 Zero-Shot & 83.8 & 85.3 & 64.8 & 96.8 & 61.2 & 37.2 / 36.0 / 30.8 & 34.7 & 62.0\\
 Few-Shot & \underline{93.6} & \underline{91.6} & \underline{70.6} & \underline{99.6} & \underline{97.6} & \underline{68.4} / \underline{80.8} / \underline{81.2} & \underline{76.8} & \underline{85.4}\\

  \rowcolor{gray!20}
 \multicolumn{9}{l}{\textit{Meta-Reasoning Paradigm (Ours)}} \\
 Few-Shot & {\bf 96.7} & {\bf 95.4} & {\bf 91.6} & {\bf 100.0} & {\bf 100.0} & {\bf 100.0 / 100.0 / 100.0} & {\bf 100.0} & \textbf{97.9}\\

 \midrule
 $\Delta$ & \textcolor{blue}{+3.1} & \textcolor{blue}{+3.8} & \textcolor{blue}{+21.0} & \textcolor{blue}{+0.4} & \textcolor{blue}{+2.4} & \textcolor{blue}{+31.6} / \textcolor{blue}{+19.2} / \textcolor{blue}{+18.8} & \textcolor{blue}{+23.2} & \textcolor{blue}{+12.5}\\
 \midrule

 \multicolumn{9}{c}{\bf ChatGPT (\texttt{GPT-3.5-Turbo})} \\

\midrule
 
\rowcolor{gray!20}
 \multicolumn{9}{l}{\textit{Chain-of-Thought Paradigm}} \\
 Zero-Shot & 91.5 & 85.5 & 75.6 & 96.4 & 68.8 & 55.6 / 54.0 / 43.2 & 50.9 & 71.3 \\
 Few-Shot & \underline{95.2} & \underline{93.9} & \underline{80.2} & \underline{99.2} & \underline{96.0} & \underline{62.8} / \underline{57.2} / \underline{54.0} & \underline{58.0} & \underline{79.8} \\

  \rowcolor{gray!20}
 \multicolumn{9}{l}{\textit{Meta-Reasoning Paradigm (Ours)}} \\
 Few-Shot & {\bf 98.7} & {\bf 98.0} & {\bf 92.4} & {\bf 100.0} & {\bf 99.2} &  {\bf 100.0 / 88.0 / 84.4} & {\bf 90.8} & {\bf 95.1} \\

 \midrule
 $\Delta$ & \textcolor{blue}{+3.5} & \textcolor{blue}{+4.1} & \textcolor{blue}{+12.2} & \textcolor{blue}{+0.8} & \textcolor{blue}{+3.2} & \textcolor{blue}{+37.2} / \textcolor{blue}{+30.8} / \textcolor{blue}{+30.4} & \textcolor{blue}{+32.8} & \textcolor{blue}{+15.3} \\
 \midrule

\end{tabular}
}

\end{center}
\vspace{-0.1in}
\caption{{\bf Conventional Reasoning Results:} We apply our method on 175B GPT-3 (text-davinci-002 and -003) and ChatGPT, and compare it with three common paradigms: Fine-tuned, Standard Prompting, and Chain-of-Thought Prompting. Our performance gains ($\Delta$) are computed over the previous SOTA (\underline{underline}). Track(\textit{Avg.}) represents the averaged accuracy of Track(3/5/7), and \textbf{Avg.} represents the average accuracy across all datasets.}
\label{tab:mainresults}
\end{table*}

\subsection{Setup}
\label{subsec:setup}

\paragraph{Tasks and Datasets.}
We conduct experiments on two categories: 
(i) conventional reasoning, involving basic reasoning scenarios like arithmetic, symbolic, and logical reasoning. This includes the following datasets: MultiArith \citep{roy2015solving}, AddSub \citep{hosseini2014learning}, Last Letter Concatenation (Letter) \citep{weichain}, Coin Flip (Coin) \citep{weichain}, Web of Lies (Lies) \citep{srivastava2022beyond}, Tracking Shuffled Objects\footnote{Divided into 3 subsets based on the number of objects and shuffler operations (3/5/7).} (Track) \citep{srivastava2022beyond},
and (ii) interactive reasoning, which involves reasoning scenarios of multi-agent mental gaming, including Hi-ToM\footnote{Divided into 5 subsets based on the number of mental gaming orders (1/2/3/4/5).} \citep{he2023hi}.
Refer to Appendix \ref{sec:dataset_details} for detailed information on datasets.

\paragraph{Language Models.}
We utilize publicly available 175B GPT-3 models (\texttt{text-davinci-002} and \texttt{text-davinci-003}) \citep{brown2020language}, as well as ChatGPT (\texttt{gpt-3.5-turbo}).\footnote{\url{https://chat.openai.com/}} 
Additionally, for comparison purposes, we include other robust closed-API LLMs: 175B Codex (\texttt{code-davinci-002}) \citep{chen2021evaluating} and 540B PaLM \citep{chowdhery2022palm}.

\paragraph{Implementation and Baselines.}
In our Meta-Reasoning (MR) paradigm, we use the completely-serial mode for arithmetic tasks and the crossly-serial mode for symbolic and logical tasks.
We also compare our method with three other paradigms: (i) Fine-tuning; (ii) Standard prompting, including Zero-Shot and Few-Shot; (iii) Chain-of-Thought (CoT) prompting, including Zero-Shot-CoT \citep{kojima2022large} and Few-Shot-CoT \citep{weichain}.
Refer to Appendix \ref{sec:demonstration} for demonstrations.

\subsection{Main Results I: Conventional Reasoning}
\label{sec:basic_results}

\paragraph{Overall Performances.}
Table \ref{tab:mainresults} presents the results.\footnote{Experimental results of GPT-3 were obtained in March 2023 via the OpenAI API interface, while the results of ChatGPT were obtained in November 2023.}
Our MR consistently outperforms Few-Shot-CoT and notably excels on complex tasks challenging for LLMs.
This trend is particularly evident for the relatively capacity-constrained text-davinci-002.
Notably, on intricate tasks where pure CoT struggles, our MR effectively alleviates the reasoning bottleneck, resulting in significantly higher accuracy (+27.0\% in Letter and +37.7\% in Track).
This indicates that our MR facilitates LLMs in learning general principles for specific task types, automatically reducing reasoning difficulty across various semantic representations.

\begin{figure*}[t]
  \centering
  \includegraphics[width=1\textwidth]{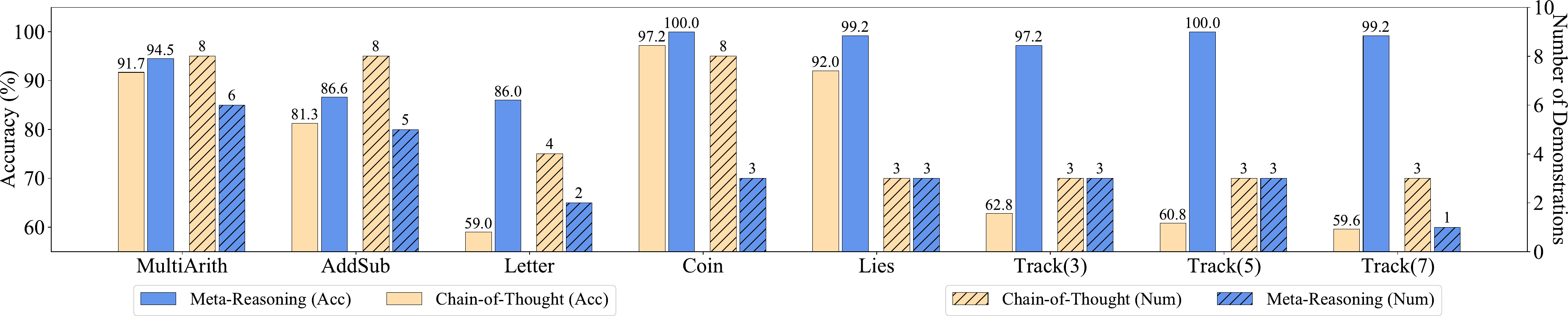}  
  \vspace{-0.2in}
  \caption{The number of demonstrations used in the CoT and MR paradigms and the corresponding performances.}
  \label{img:domenstrations}
\end{figure*}

\paragraph{Fewer Demonstrations, Better Performances.}
\label{subsec:efficient}

Figure \ref{img:domenstrations} shows comparisons between the performance of CoT and MR paradigms with varying numbers of demonstrations.
MR consistently achieves superior performance across almost all datasets while utilizing fewer demonstrations, particularly evident in symbolic and logical reasoning tasks.
For example, in the Letter task, MR results in a +27.0\% improvement for LLMs with 1/2 demonstrations compared to the CoT paradigm. Similarly, in the Track(7) task, using only 1/3 demonstrations (i.e., one demonstration) leads to a remarkable +39.6\% boost.
This indicates that LLMs can acquire general solutions for specific tasks with minimal demonstrations, facilitating learning through analogy.

\begin{figure*}[htb]
  \centering
  \includegraphics[width=0.96\textwidth]{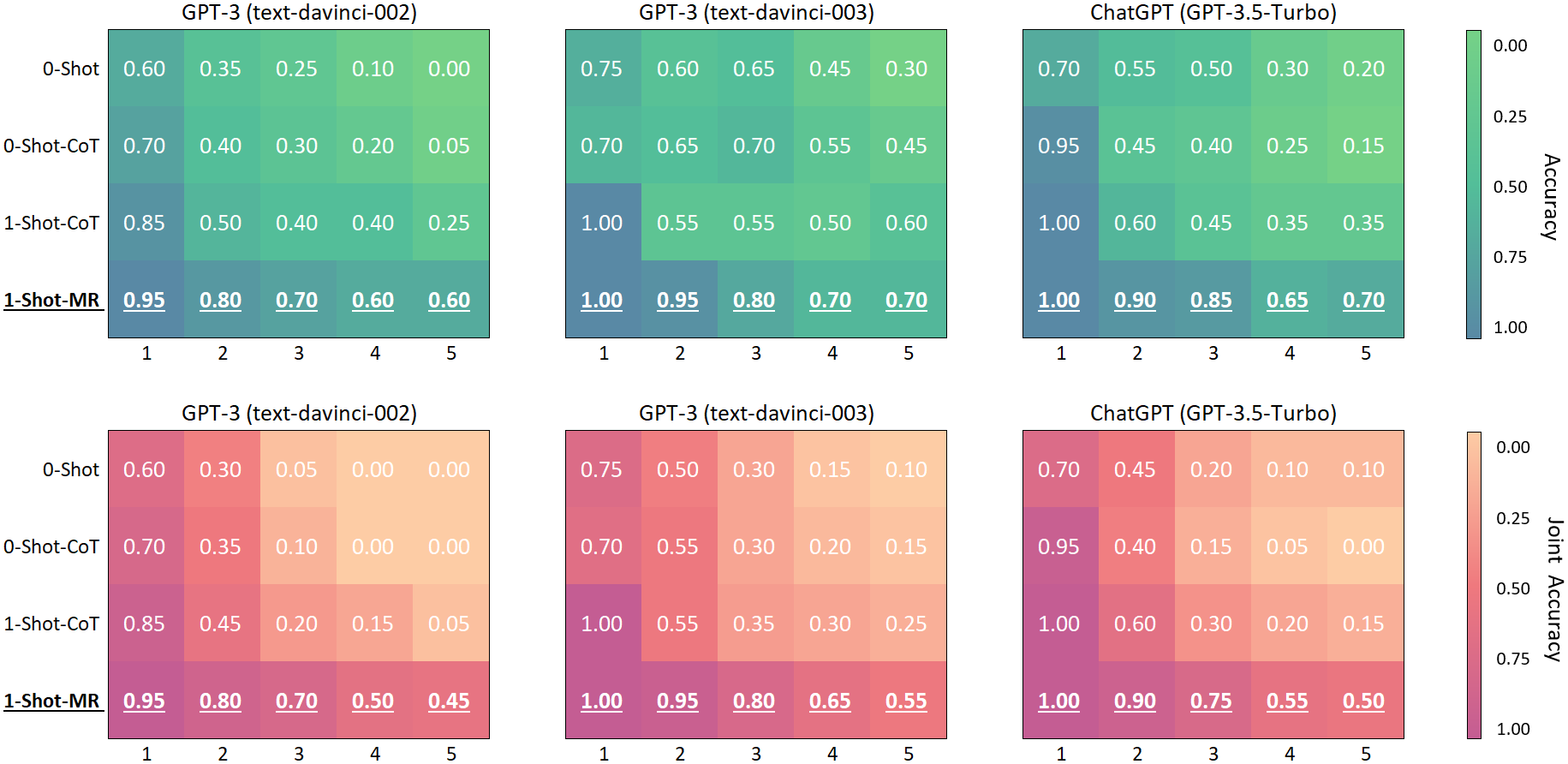}  
  \vspace{-0.0in}
  \caption{{\bf Interactive Reasoning Results:} The accuracy (upper part) and joint accuracy (lower part) of GPT-3 (text-davinci-002 and -003) and ChatGPT on the Hi-ToM dataset. The $x$-axis of each heatmap represents ToM orders. [Metric Explanation: (i) \textit{Accuracy} refers to the correctness of each order independently. (ii) \textit{Joint Accuracy} reflects the cumulative correctness, wherein the $k$-order reasoning is deemed correct only if all reasoning orders less than $k$ are also correct. This metric is instrumental in mitigating randomness error.]}
  \label{img:HiToM}
\end{figure*}

\subsection{Main Results II: Interactive Reasoning}
\label{sec:advanced_results}

The real-world reasoning environment is more intricate than these conventional reasoning scenarios.
Therefore, we consider more complex interactive scenarios and introduce the Theory-of-Mind (ToM) reasoning.
In ToM reasoning, the objects involved in reasoning require subjective observation or cognitive abilities, and their observation and thought directly influence the reasoning outcomes.
Therefore, LLMs are susceptible to interference.

The variable parameter ``Order'' determines ToM's difficulty level, which refers to the layer number of the mental game involved.
For example, in 3-order reasoning, the structure might be ``\textit{A thinks B thinks C thinks xxx}''.
Notably, 1-order reasoning does not entail any interaction and is categorized as low-order reasoning. On the other hand, reasoning with an order greater than 1 involves a mental game between multiple observers and is classified as high-order reasoning.

When solving lower-order ToM questions, both 1-shot CoT and MR achieve nearly 100\% accuracy, indicating that LLMs can accurately comprehend the reasoning text itself.
But when solving high-order ToM, CoT exhibits a notable performance decline, with an about 40\% decrease in joint accuracy when transitioning from 1 to 2-order, and with a nearly 0\% accuracy remaining at 5-order.
In contrast, MR maintains stable performance as the order increases. At 5-order, its performances equal 2-order performances of CoT, indicating its strong ability to handle complex reasoning.

\section{Advantage Analysis}

\subsection{Boundary Test: OOD Generalization}
\label{sec:out-of-domain_results}

Out-of-domain (OOD) generalization highlights LLMs' ability to address novel tasks by synthesizing limited in-domain knowledge \citep{wang2024trajectory}.
We set a challenging boundary test involving Lies, Track, and ToM tasks, to compare the OOD boundary of MR and CoT methods.

For each task, we first manually dissect the smallest unit of reasoning (Details are shown in Appendix \ref{sec:reasoning_unit}). Within each demonstration, we limit the reasoning units to three; thus, any new question exceeding this threshold is considered OOD.
We generate 50 samples per task without any reasoning units, then progressively incorporate reasoning units adhering to the structure of the respective dataset.
When the following situation occurs for the first time: when the sample contains $k$ reasoning units, LLMs answer correctly; when it contains $k+1$ reasoning units, LLMs answer incorrectly. At this point, the sample stops iterating, and its Boundary Length (${\rm BL}$) is recorded as $k$.
The sample iteration ceases upon encountering the first case where LLMs answer accurately with $k$ reasoning units and inaccurately with $k+1$ reasoning units. The Boundary Length of this sample\footnote{Refer to Appendix \ref{sec:algorithm} for a detailed algorithm process.} is recorded as $k$.
In dataset $\mathcal{D}$, we compute the Boundary Rate (${\rm BRate}$) for each $k \leq k_{\rm max}$ as the following formulation:
\begin{equation}
    {\rm BRate}(\mathcal{D}, k) = \frac{\sum_{s \sim \mathcal{D}} \mathbb{I}({\rm BL}(s) \geq k)}{|\mathcal{D}|},
\end{equation}
where $\mathbb{I}(\cdot)$ is the indicator function, $k_{\rm max}$ is the maximum number of reasoning units.

\begin{figure}[t]
  \centering
  \includegraphics[width=0.45\textwidth]{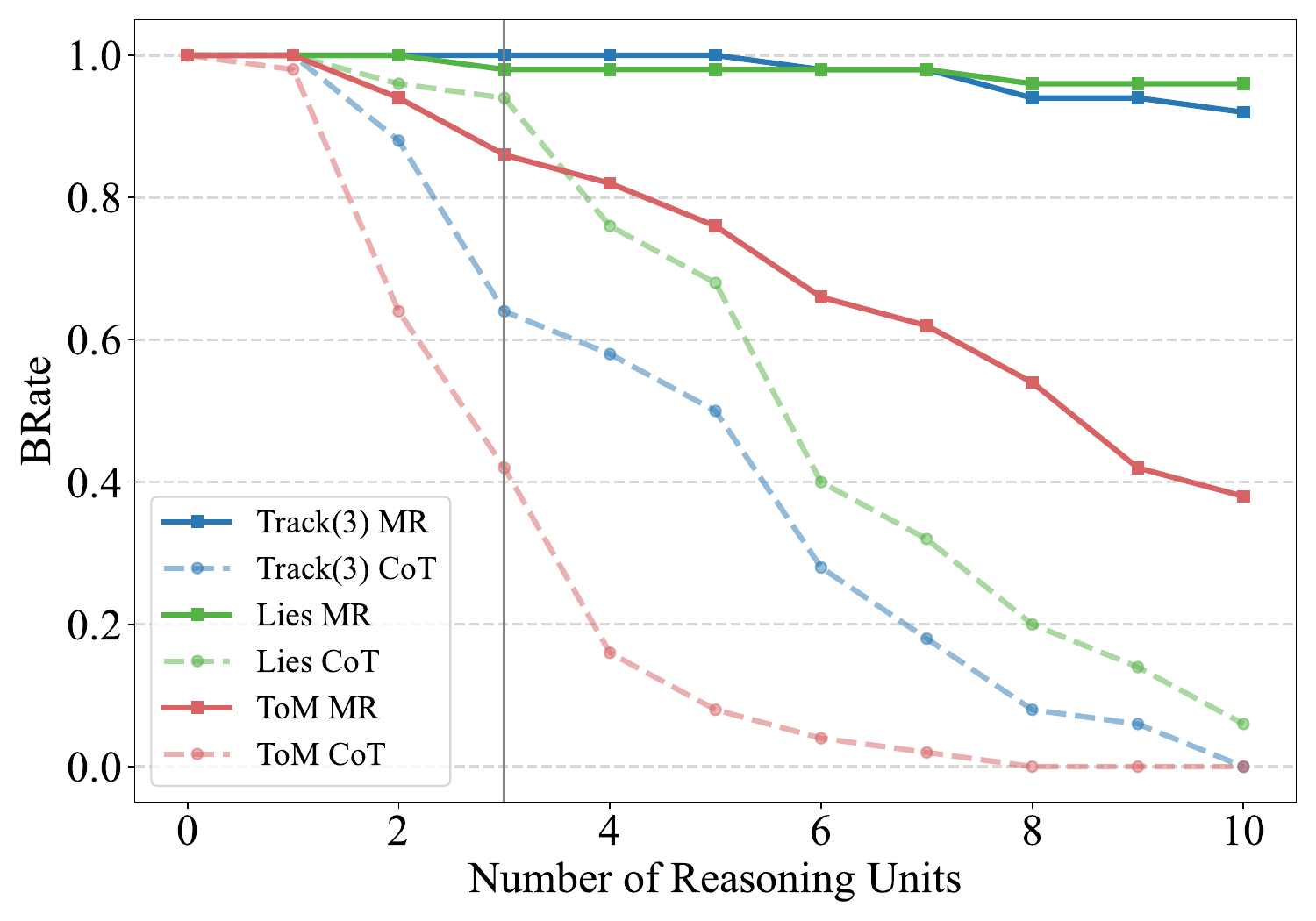}  
  \vspace{-0.03in}
  \caption{Boundary test of out-of-domain generalization under CoT and MR paradigms, where the number of reasoning units is larger than three (the right area of the vertical gray line in the figure) means out-of-domain.}
  \label{img:boundary_test}
\end{figure}

We draw ${\rm BRate}$ curves of each $\mathcal{D}$. The larger the area enclosed by the curve and the $x$-axis, the stronger the OOD generalization of the method.
Figure \ref{img:boundary_test} shows the ${\rm BRate}$ curves of each dataset under CoT and MR paradigms, respectively.
We note that as the number of reasoning units grows beyond the domain, the CoT curves exhibit a sharp decline, while the MR curves maintain relative smoothness, with Lies and Track tasks achieving nearly 100\% ${\rm BRate}$.
This indicates that our MR facilitates strong OOD generalization for LLMs.

\begin{figure}[t]
  \centering
  \includegraphics[width=0.9\columnwidth]{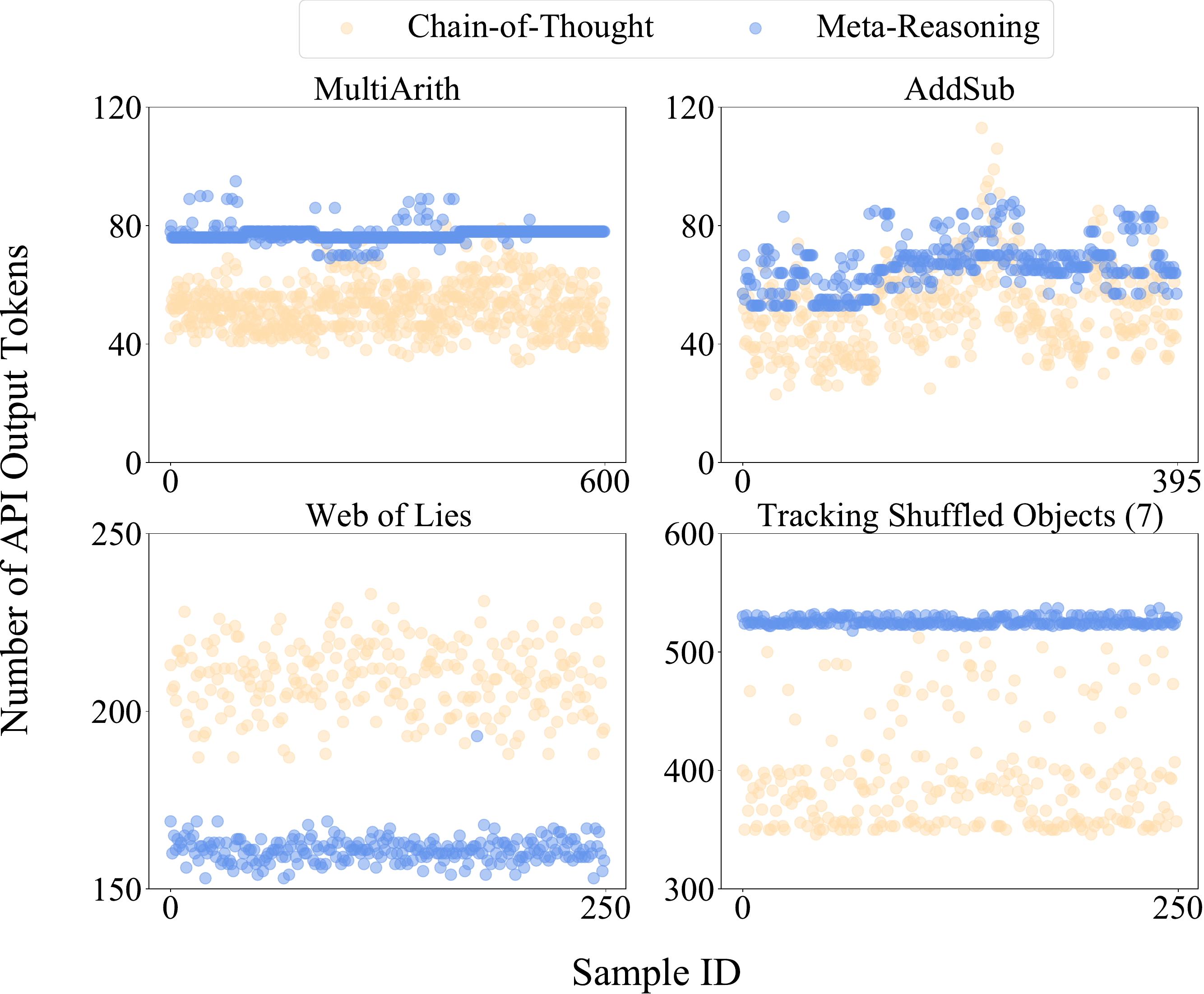}  
  \vspace{-0.03in}
  \caption{The token number distributions of the text generated by GPT-3 text-davinci-002 when using the Chain-of-Thought and Meta-Reasoning paradigms.}
  \label{img:api}
\end{figure}

\subsection{Output Stability Test}

In addition to performance, user experience is another crucial consideration.
Currently, access to LLMs like GPT-3 involves paywalls. Unexpected outputs, such as endless looping text or random guessing, can increase user fees, so making a stable output space is essential.
We analyze the number of API output tokens generated by MR and CoT paradigms for each sample to evaluate this stability, as illustrated in Figure \ref{img:api}.
When employing the MR, the output scales of different samples are much closer. Conversely, under the CoT, outputs scatter widely, increasing the likelihood of encountering unexpected and abnormal situations.

\section{Discussion}

We conduct ablation studies to examine the role of semantic resolution in the reasoning process.
Moreover, we compare our method with existing work in language programming \cite{chen2022program,gao2023pal}, highlighting Meta-Reasoning's broader applicability across diverse scenarios.
These extended analyses are shown in Appendix \ref{sec:analysis}.

\section{Related Work}
\label{sec:related_work}
Our work is related to the research lines of neural-symbolic methods and chain-of-thought reasoning. Please refer to Appendix \ref{sec:additional_work} for full details.

\paragraph{Neural-Symbolic Methods in LLMs.}
Symbolic learning \citep{chen2021evaluating} significantly improves LLMs' reasoning performance. Prior works focus on converting natural languages into programming languages \citep{gao2022pal,cheng2022binding} and accessing external interpreters for execution \citep{schick2023toolformer}; or using symbolic tasks for post-tuning \citep{liu2023zero, wei2023symbol}, leading to performance improvements.
However, these symbols are well-defined formal languages completely independent of natural languages.
Our work jumps out of this framework and further enhances the efficiency of the symbolic methods.

\paragraph{Chain-of-Thought Reasoning.}
Intriguing chain-of-thought techniques \citep{weichain,kojimalarge,wang2022self,zhang2022automatic} have effectively leveraged the emergent ability of LLMs to decompose multi-step reasoning.
It can improve the performance of general-purpose and even domain-specific reasoning \citep{zhang2023multicot,wang2023element,he2023exploring,zhang2023igniting}.

\section{Conclusion}
We propose Meta-Reasoning, a semantic-symbol deconstruction paradigm for reasoning. Through the semantic resolution of the original questions, we enable LLMs to grasp meta-forms and general solutions for specific types of reasoning tasks.
This approach requires fewer demonstrations to expand the upper limit of their reasoning accuracy, out-of-domain generalization, and output stability.

\section*{Limitations}
Semantic resolution dictates that Meta-Reasoning tasks must disregard the intrinsic properties of entities. Consequently, Meta-Reasoning may not be well-suited for reasoning tasks reliant on world knowledge in semantics, such as commonsense reasoning. However, Meta-Reasoning shows potential in real-world agent reasoning scenarios \citep{gao2023jsontuning,tang2023struc}. When agents are impeded by irrelevant properties, Meta-Reasoning can effectively circumvent such obstacles. We aim to explore more comprehensive reasoning scenarios to further justify its applicability in future work.

\section*{Ethics Statement}
We use publicly available datasets for experiments, so the ethics issues of the source texts are non-existent.
For the generated contents with LLMs, prior work \citep{brown2020language, chan2023gpt} has elaborated on their inevitable potential toxicity, such as issues of bias and fairness.
We completely keep the prompts neutral and task-specific to avoid toxic language generation, and there were no toxic texts that appeared in our experiments.

\section*{Acknowledgements}
Yiming and Rui are with MT-Lab, Department of Computer Science and Engineering, School of Electronic Information and Electrical Engineering, and also with the MoE Key Lab of Artificial Intelligence, AI Institute, Shanghai Jiao Tong University, Shanghai 200204, China.  This paper is mainly supported by the Alibaba-AIR Program (22088682). Yiming and Rui are also supported by the National Natural Science Foundation of China (62176153), and the Shanghai Municipal Science and Technology Major Project (2021SHZDZX0102). This work is also partially supported by the Joint Funds of the National Natural Science Foundation of China (U21B2020).


\bibliography{anthology,custom}
\appendix

\newpage

\section{Semantic-Symbol Operation Rulebase}
\label{sec:rulebase}

Table \ref{tab:ablation} shows operation mapping examples.
Due to the lack of automatic methods, the rule base is continuously revised and improved with the annotation process.

\begin{table}[htb]
\begin{center}
\small
\resizebox{0.82\columnwidth}{!}
{
\begin{tabular}{c|l}

\toprule

Symbol & Semantics \\
\midrule
= & is, are, have, ... \\
$\rightarrow$ & mean, represent, infer, ...\\
+ & buy, get, pick, ... \\
- & sell, throw, lose, ... \\
$\times$ & each, per, both, ... \\
$\div$ & split, divide, group, ... \\
\bottomrule

\end{tabular}
}
\vspace{-0.0in}
\caption{Examples of operations with infinite natural semantics mapped to finite symbols.}
\label{tab:operation}%
\end{center}
\end{table}

\section{Dataset Details}
\label{sec:dataset_details}

To measure the generalizability of our approach, we consider conventional and interactive reasoning:

\paragraph{Conventional Reasoning.} 
In this scenario, reasoning information is globally accessible to all observers. 
We adopt three categories of reasoning as our testbed: 
(i) Arithmetic reasoning, we choose MultiArith \citep{roy2015solving} and AddSub \citep{hosseini2014learning} tasks, with 600 and 395 test instances separately; 
(ii) Symbolic reasoning, we follow \citet{weichain} to use Last Letter Concatenation and Coin Flip tasks, they both include 500 test instances;
(iii) Logical reasoning, We choose Web of Lies and Tracking Shuffled Objects tasks from BIG-bench \citep{srivastava2022beyond} --- a more challenging reasoning task collection. In particular, the Tracking Shuffled Objects task is divided into three datasets according to the number of objects and shuffler operations (3/5/7). each dataset includes 250 test instances.

\paragraph{Interactive Reasoning.} 
In this scenario, individual observers are limited to observing distinct local reasoning information, necessitating reliance on interaction and mental gaming for their reasoning processes.
We select the Theory-of-Mind (ToM) task as our testbed and choose Hi-ToM \citep{he2023hi} as a benchmark for it involves the complex higher-order mind. This dataset contains a collection of multiple subsets ranging from 1 to 5 orders, each subset has 20 test instances.

\begin{figure*}
  \centering
  \includegraphics[width=\textwidth]{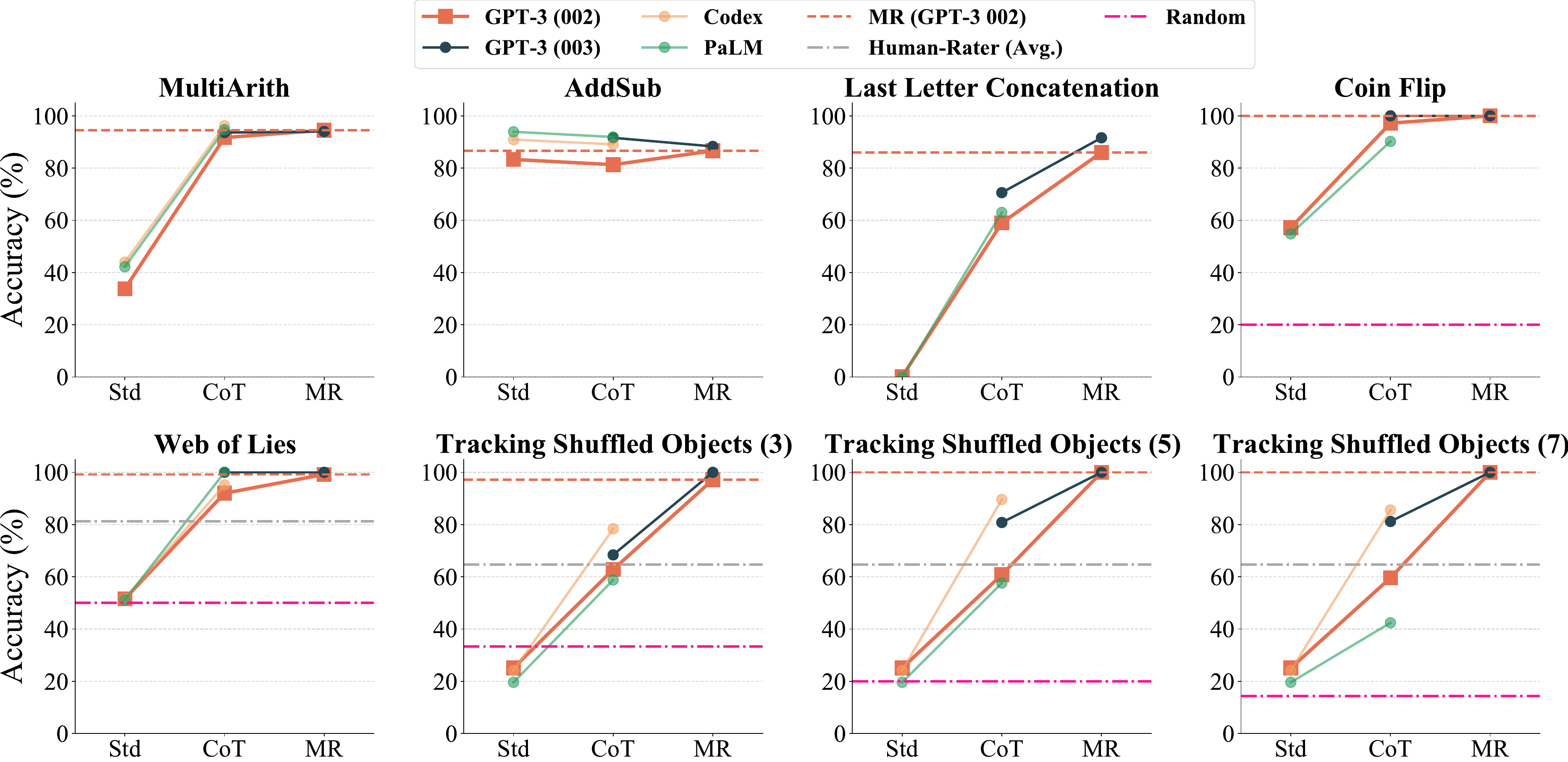}  
  \vspace{-0.25in}
  \caption{The performance gaps between four LLMs under different paradigms (Std $\rightarrow$ Standard prompting, CoT $\rightarrow$ Chain-of-Thought, MR $\rightarrow$ Meta-Reasoning) in all datasets.}
  \label{img:llmcmp}
\end{figure*}

\section{Extended Analysis}
\label{sec:analysis}

\subsection{Bridge the Gap between LLMs' Capabilities.}
We conduct longitudinal analyses of performance gaps between four LLMs.
Figure \ref{img:llmcmp} visualizes the performance gaps between LLMs for the same dataset and paradigm.
Observations are as below: 

\begin{itemize}[leftmargin=0.3cm]
    \item GPT-3 text-davinci-002 (the worst original performance among the four LLMs) greatly outperforms the remaining three LLMs under the CoT paradigm on five datasets after adopting the Meta-Reasoning paradigm.
    \item Performance gaps between text-davinci-002 and -003 on all datasets are greatly reduced compared to under the CoT paradigm after adopting the Meta-Reasoning paradigm.
\end{itemize}

These findings indicate that our Meta-Reasoning paradigm further bridges the gap in the LLMs' capability themselves, allowing the weaker LLMs (e.g. text-davinci-002) to approximate the stronger LLMs (e.g. text-davinci-003) in reasoning ability.

\subsection{Ablation Study}
\label{subsec:ablation}

We perform ablation studies to explore the role of semantic resolution in the whole reasoning process.
Table \ref{tab:ablation} reports the error rates of all datasets under both paradigms and error reason rates (caused by semantic resolution or pure reasoning) in the wrong samples for each dataset.

\begin{table*}[t]

\begin{center}
\resizebox{0.98\textwidth}{!}
{
\begin{tabular}{ll|cccccc}
\toprule

& & MultiArith & AddSub & Letter & Coin & Lies & Track(\textit{Avg.}) \\
\midrule
\multirow{2}{*}{Error Rate (\%)} & Chain-of-Thought & 8.3 & 18.7 & 41.0 & 2.8 & 8.0 & 38.9 \\
& Meta-Reasoning & 5.5 & 13.4 & 14.0 & 0.0 & 0.8 & 1.2 \\
\midrule
\multirow{2}{*}{\makecell{Error Reason Rate (\%) \\ {\footnotesize (Meta-Reasoning)}}} & Semantic Resolution & 84.8 & 67.9 & 0.0 & 0.0 & 0.0 & 0.0 \\
& Pure Reasoning & 15.2 & 32.1 & 100.0 & 0.0 & 100.0 & 100.0 \\
\bottomrule

\end{tabular}
}
\end{center}
\vspace{-0.1in}
\caption{Error rates using the Chain-of-Thought and Meta-Reasoning paradigms for all datasets, and error rates caused by semantic resolution and pure reasoning when using the Meta-Reasoning paradigm. Note that \textit{under each dataset, the error rates of semantic resolution and pure reasoning sum up to a constant 1}. This arises from the fact that when semantic resolution errors occur, we no longer classify pure reasoning as either correct or incorrect. 
For instance, within the MultiArith dataset, among the 5.5\% of error samples, 84.8\% were attributed to semantic reasoning inaccuracies, leaving the remaining 15.2\% attributed to errors in pure reasoning.}
\label{tab:ablation}
\end{table*}

We note that the causes of errors are inconsistent in different reasoning scenarios.
For symbolic and logical reasoning, LLMs hardly produce any semantic resolution errors, only errors in the reasoning process (of course, the error rate of their reasoning itself is extremely low). This shows that semantic reasoning fully plays a positive role in reducing the complexity of reasoning for LLMs.
But in arithmetic reasoning, semantic resolution errors often occur, and exceed the errors in the reasoning process itself. This shows that LLMs cannot reduce all types of questions under specific arithmetic datasets well.
Intuitively, symbolic and logical reasoning questions are easier to logicalize than arithmetic reasoning questions, and the combination of reasoning units under arithmetic reasoning is more flexible. How to fully push the upper limit of LLM's semantic resolution ability, so as to further improve its reasoning ability, is a promising future work.

\subsection{Formal Pattern Flexibility}

\begin{figure}[t]
  \centering
  \includegraphics[width=0.95\columnwidth]{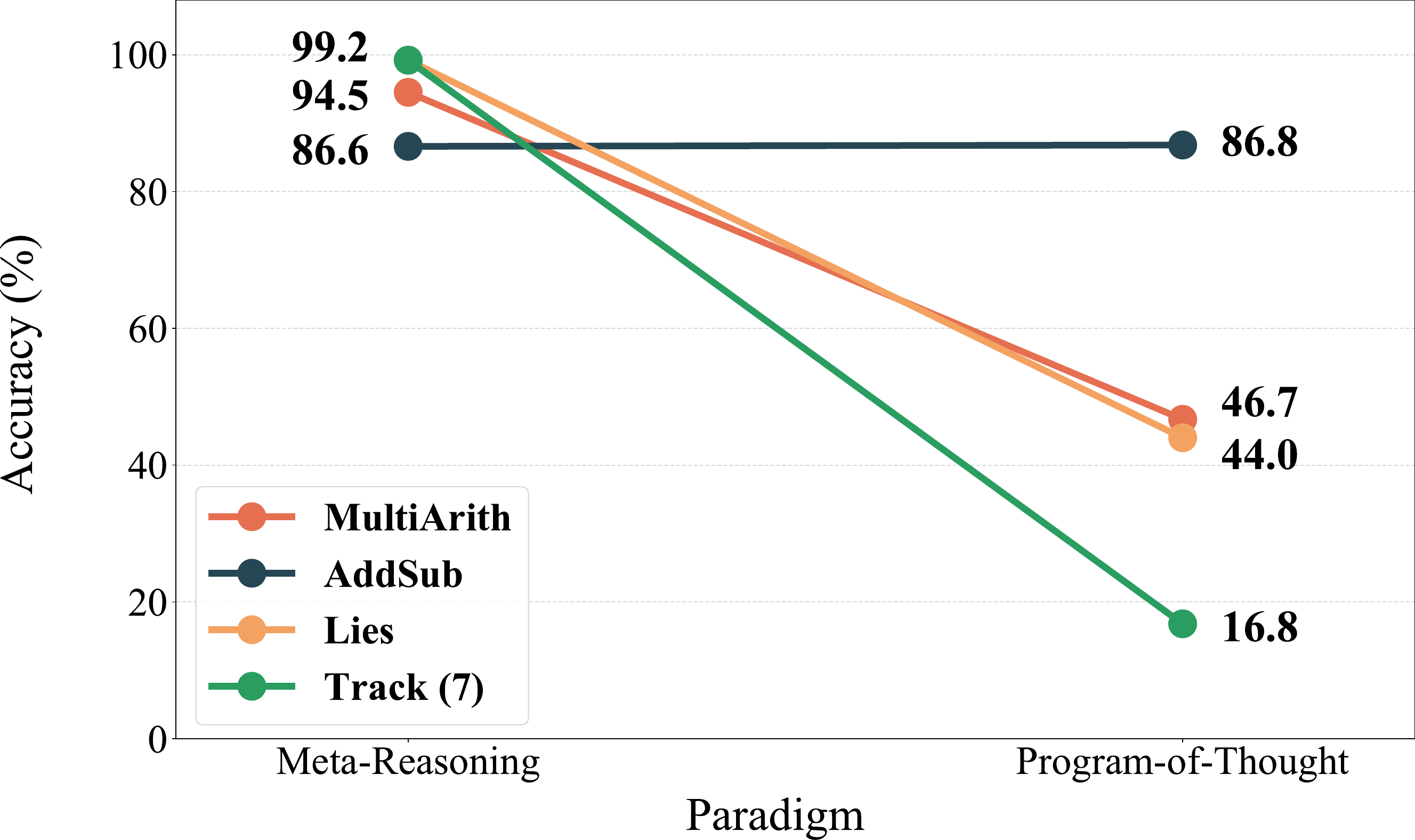}
  \vspace{-0.1in}
  \caption{Performance comparisons between our Meta-Reasoning paradigm and the Program-of-Thought paradigm (w/o external Python interpreter).}
  \label{img:PoT}
\end{figure}

So far, most symbolic reasoning work focuses on mapping natural semantics to formal languages with complete grammar (such as Python and SQL).
However, this grammatical completeness actually limits the form conversion, and it has higher requirements for the abstraction of the original reasoning tasks.
To verify the flexibility of our paradigm, we contrast Program-of-Thought (PoT), a Text-to-Python reduction approach for reasoning tasks \cite{chen2022program,gao2022pal}.
Meanwhile, to keep the settings consistent, we eliminate the call of an external interpreter in the PoT paradigm but utilize the LLMs themselves to complete the entire reasoning step, and select the same demonstrations for the two paradigms.

Figure \ref{img:PoT} shows the performance comparisons of PoT and MR paradigms on four datasets.
For the two arithmetic reasoning tasks (MultiArith and AddSub), the performance of PoT fluctuates wildly after removing the external interpreter.
For the two symbolic reasoning tasks (Lies and Track), PoT is almost completely ineffective.
In contrast, MR has stronger flexibility when encountering reasoning tasks that are not easily programmed.

\begin{algorithm}[t]
    \renewcommand{\algorithmicrequire}{\textbf{Input:}}
    \renewcommand{\algorithmicensure}{\textbf{Output:}}
    \caption{Computation of Boundary Length}
    \label{alg1}
    \begin{algorithmic}[1]
        \REQUIRE  
        $x$: Initialized sample w/o reasoning units.\\
        $\mathcal{D}$: Source dataset of $x$.\\
        $g(\mathcal{D})$: Reasoning unit generator imitating the style of $\mathcal{D}$.\\
        $k_{\rm max}$: Maximum number of reasoning units.\\
        $p_{\theta}$: Language Model.

        \STATE $k \leftarrow 0$
        \WHILE{$k < k_{\rm max}$}
            \STATE $u \rightarrow g(\mathcal{D})$, $x \leftarrow x + u$, $y \leftarrow p_{\theta}(x)$
            \IF{$y$ is the correct answer of $x$}
                \STATE $k \leftarrow k + 1$
            \ELSE \STATE break
            \ENDIF
        \ENDWHILE
        \RETURN $k$
    \end{algorithmic}  
\label{algorithm}
\end{algorithm}

\section{Details of Boundary Test}

\subsection{Reasoning Unit Division}
\label{sec:reasoning_unit}

Sample reasoning units for three datasets are as below. The smallest reasoning unit is highlighted in \colorbox{cyan}{blue}.

\begin{itemize}[leftmargin=0.5cm]
    \item \textbf{Lies.} \\
    Andree lies. \\
    \colorbox{cyan}{Delfina says Andree lies}. \\ 
    \colorbox{cyan}{Jim says Delfina tells the truth}. \\ \colorbox{cyan}{Gwenn says Jim lies}. \\ 
    \colorbox{cyan}{Delbert says Gwenn lies}. \\
    Does Delbert tell the truth?
    \item \textbf{Track.}  \\
    Alice, Bob, and Claire are dancers at a square dance. At the start of a song, they each have a partner: Alice is dancing with Lola, Bob is dancing with Patrick, and Claire is dancing with Melissa. Throughout the song, the dancers often trade partners. \\
    First, \colorbox{cyan}{Alice and Claire switch partners}. \\
    Then, \colorbox{cyan}{Bob and Claire switch partners}. \\
    Finally, \colorbox{cyan}{Claire and Alice switch partners}. \\
    At the end of the dance, Bob is dancing with
    \item \textbf{ToM.} \\
    \{\texttt{Scenario}\} \\
    Where does \colorbox{cyan}{Isabella think} \colorbox{cyan}{Owen thinks} \colorbox{cyan}{Charlotte thinks} \colorbox{cyan}{Aver thinks} the lettuce is?
\end{itemize}

\subsection{Computation of Boundary Length}
\label{sec:algorithm}

The algorithm of Boundary Length (BL) Computation is shown in Algorithm \ref{algorithm}.

\section{Case Study}

Table \ref{tab:cases} compares output examples under Few-Shot-CoT and Meta-Reasoning paradigms.
We find that when the original question requires reasoning about a large number of entities and the corresponding relationships, the reasoning process in the Few-Shot-CoT paradigm tends to be more chaotic, leading to errors and potential confusion between entities.
In contrast, Meta-Reasoning offers a clearer and more concise reasoning process, reducing interference caused by a large number of natural semantics.

\begin{table*}[htb]

\centering
\scriptsize
  \renewcommand\arraystretch{1}
  \setlength{\tabcolsep}{2.5mm}{

\begin{tabular}{p{0.95\textwidth}}

\toprule
\textbf{Question (Tracking Shuffled Objects)} \\
\midrule

Q: Alice, Bob, and Claire are dancers at a square dance. At the start of a song, they each have a partner: Alice is dancing with Lola, Bob is dancing with Rodrigo, and Claire is dancing with Patrick.\\
Throughout the song, the dancers often trade partners. First, Alice and Bob switch partners. Then, Claire and Bob switch partners. Finally, Bob and Alice switch partners. At the end of the dance, Alice is dancing with\\
Options:\\
(A) Lola\\
(B) Rodrigo\\
(C) Pa\\

\midrule
\textbf{Few-Shot-CoT} \\
\midrule

\textit{A: Let's think step by step.} (0) At the start: Alice: Lola, Bob: Rodrigo, Claire: Patrick.\\
(1) Alice and Bob switch partners: Alice: Rodrigo, Bob: Lola, Claire: Patrick.\\
(2) Claire and Bob switch partners: Alice: Rodrigo, Bob: Patrick, Claire: Lola.\\
\textcolor{red}{(3) Bob and Alice switch partners: Alice: Lola, Bob: Patrick, Claire: Rodrigo.}\\
At the end of the dance, Alice is dancing with Lola. So the answer is (A).
{\color{red}{\XSolidBrush (Entity information interferes with reasoning)}}\\

\midrule
\textbf{Meta-Reasoning}\\
\midrule

A: The question can be simplified to: It is known that A = 1, B = 2, C = 3.\\
First, B and A swap books: B and A $\rightarrow$ (B = 2, A = 1 $\rightarrow$ B = 1, A = 2) $\rightarrow$ A = 2, B = 1, C = 3.\\
Then, A and C swap books: A and C $\rightarrow$ (A = 2, C = 3 $\rightarrow$ A = 3, C = 2) $\rightarrow$ A = 3, B = 1, C = 2.\\
Finally, C and B swap books: C and B $\rightarrow$ (C = 2, B = 1 $\rightarrow$ C = 1, B = 2) $\rightarrow$ A = 3, B = 2, C = 1.\\
At the end of the semester, Alice has: Alice $\rightarrow$ A = 3, 3 $\rightarrow$ the 3-rd option $\rightarrow$ the answer is (C). {\color{blue}{\checkmark}}
\\
\toprule

\toprule
\textbf{Question (Web of Lies)} \\
\midrule

Q: Sherrie tells the truth. Ryan says Sherrie lies. Bernita says Ryan tells the truth. Tamika says Bernita lies. Jerry says Tamika lies. Does Jerry tell the truth?\\

\midrule
\textbf{Few-Shot-CoT} \\
\midrule

\textit{A: Let's think step by step.} (1) Sherrie tells the truth. So, we know that Sherrie tells the truth.\\
(2) Ryan says Sherrie lies. Since we know from (1) that Sherrie tells the truth, if Ryan says Sherrie lies, then Ryan lies.\\
(3) Bernita says Ryan tells the truth. Since we know from (2) that Ryan lies, if Bernita says Ryan tells the truth, then Bernita lies.\\
(4) Tamika says Bernita lies. Since we know from (3) that Bernita lies, if Tamika says Bernita lies, then Tamika tells the truth.\\
(5) Jerry says Tamika lies. Since we know from (4) that Tamika tells the truth, if Jerry says Tamika lies, then Jerry tells the truth.\\
Now, the question asks: Does Jerry tell the truth? We know from (5) that Jerry tells the truth. So the answer is Yes.
{\color{red}{\XSolidBrush (Complex entity relationships and long reasoning chains interfere with reasoning)}}\\

\midrule
\textbf{Meta-Reasoning} \\
\midrule

A: The question can be simplified to: It is known that A = 1.\\
Ryan says Sherrie lies: lies $\rightarrow$ A' = 0. Since A = 1, A is not equal to A', so B = 0.\\
Bernita says Ryan tells the truth: truth $\rightarrow$ B' = 1. Since B = 0, B is not equal to B', so C = 0. \\
Tamika says Bernita lies: lies $\rightarrow$ C' = 0. Since C = 0, C is equal to C', so D = 1.\\
Jerry says Tamika lies: lies $\rightarrow$ D' = 0. Since D = 1, D is not equal to D', so E = 0.\\
Since E = 0, so the answer is: no. {\color{blue}{\checkmark}}
\\

\bottomrule
\end{tabular}
}

\vspace{-0.0in}
\caption{Case studies of reasoning chains under Few-Shot-CoT and Meta-Reasoning paradigms.}
\label{tab:cases}%

\end{table*}

\section{Additional Related Work}
\label{sec:additional_work}

This work is inspired by two directions.
First is the neural-symbolic methods, which have shown great promise in improving LLMs’ reasoning performance. 
Current work mainly focuses on converting natural languages into programming languages, however, the symbols that most of these works focus on artificially defined formal languages completely independent of natural languages, which makes it hard to establish the mapping facing complex real-world scenarios. 
Therefore, our research concentrates on human natural language, delving into semantic resolution at the semiotic level, and pushing the boundaries of LLMs in handling problems within the realm of natural language.
Second is the Chain-of-Thought, an important technique for in-context learning reasoning. 
However, in-context learning with CoT is limited to learning from the reasoning process of the sample itself. Our optimization is high-level, and 
We hope to promote the efficiency and generality of sample learning by generalizing the features of a single sample to the general features of the entire dataset.
Our objective is to enhance the efficiency and generalizability of sample learning upon the CoT framework.

\paragraph{Neural-Symbolic Methods in LLMs.}
Starting from Codex \citep{chen2021evaluating}, 
 symbolic learning has shown great promise in improving LLMs' reasoning performance. Afterward, a series of works further explored symbolic approaches in LLMs' reasoning, and they can be broadly classified into two categories: (i) converting natural languages into programming languages \citep{chen2022program,gao2022pal,cheng2022binding}, such as Python or SQL, and using the powerful code capabilities of LLMs to parse and even access external interpreters for execution \citep{schick2023toolformer}; (ii) using symbolic tasks for post-tuning of LLMs \citep{liu2023zero}, which was found to lead to unexpected improvements in the overall performance of the models.
However, the ``symbols'' that most of these works focus on are artificially defined formal languages completely independent of natural languages. These works establish sample-specific one-to-one mappings between two languages (natural language $\rightarrow$ formal language).
Obviously, formal languages are learned by LLMs with less ambiguity due to their syntactic rigor, but they are divorced from the study of human natural language itself.
Recently, \citet{wei2023symbol} design a novel symbol tuning scheme by replacing natural language labels with semantically-unrelated symbols, but the symbol system they define is not complete.
This approach is different from the symbols under formal languages used in previous studies but has not been explored further in depth.
Our work closely focuses on human natural language, resolute the semantics at the semiotic level, and explores the upper limit of LLM reasoning in dealing with problems under natural language.

\paragraph{Chain-of-Thought Prompt for Reasoning.}
Intriguing chain-of-thought (CoT) techniques have effectively leveraged the emergent ability of LLMs to decompose multi-step reasoning.
Recent work in this field can be broadly classified into four categories: (i) Improving the performance of general-purpose reasoning tasks \citep{weichain,kojimalarge,wang2022self,zhou2022least,zhang2022automatic,fu2022complexity}, i.e., arithmetic, symbolic, logical, and common-sense reasoning;
(ii) Applying to domain-specific reasoning, such as multi-modality \citep{zhang2023multicot}, or some purely linguistic tasks, such as translation \citep{he2023exploring}, summarization \citep{wang2023element}, sentiment analysis \citep{fei2023reasoning}, question-answer \citep{li2022self}, etc;
(iii) Analyzing the mechanics and interpretability of CoT \citep{wang2022towards,shi2023large,lyu2023faithful};
(iv) Distilling CoT techniques for smaller models \citep{ho2022large,kim2023cot}.

\section{Demonstration Design}
\label{sec:demonstration}

\begin{figure*}
  \centering
  \includegraphics[width=\textwidth]{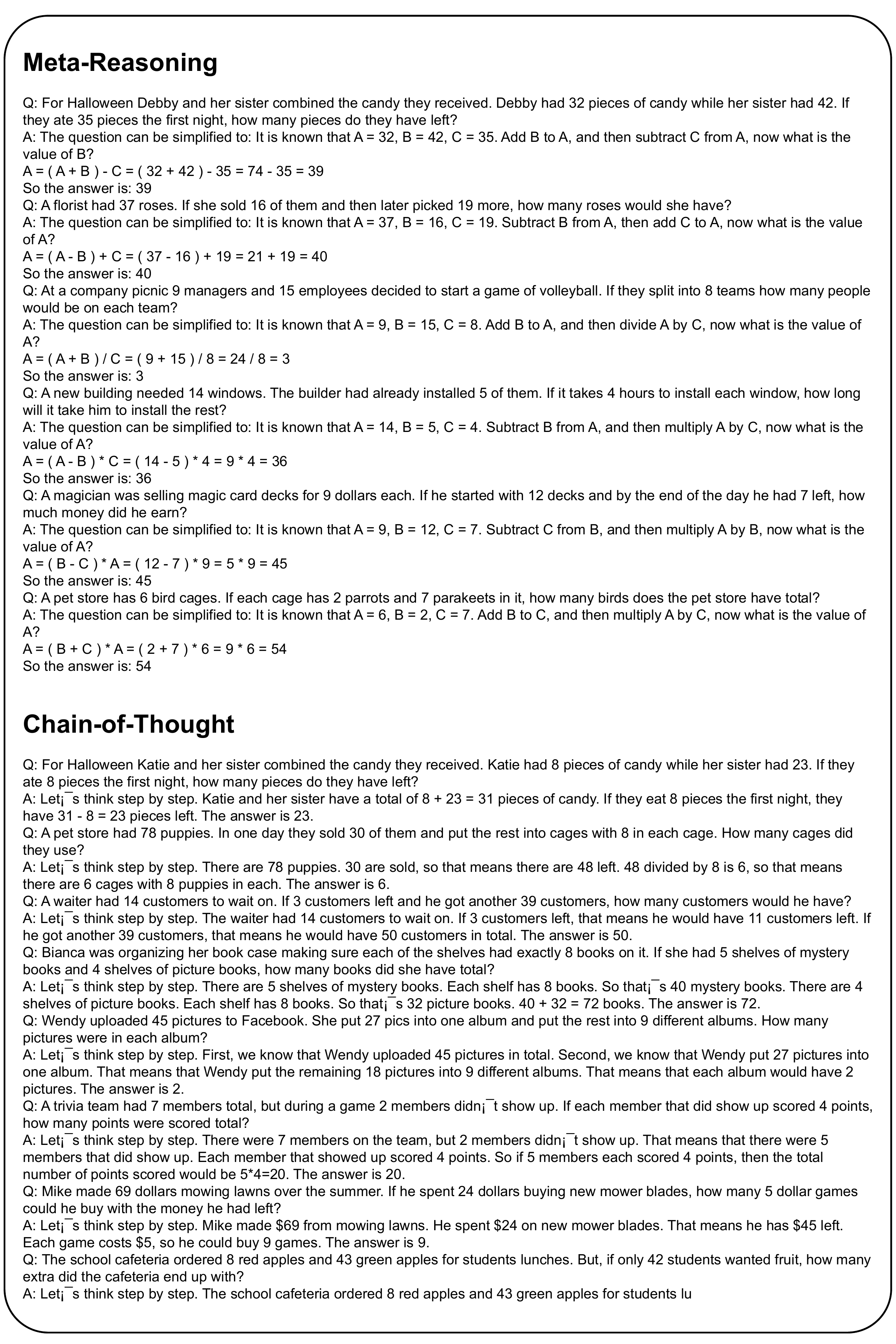}  
  \vspace{-0.25in}
  \caption{Demos: MultiArith.}
  \label{Demos: MultiArith}
\end{figure*}

\begin{figure*}
  \centering
  \includegraphics[width=\textwidth]{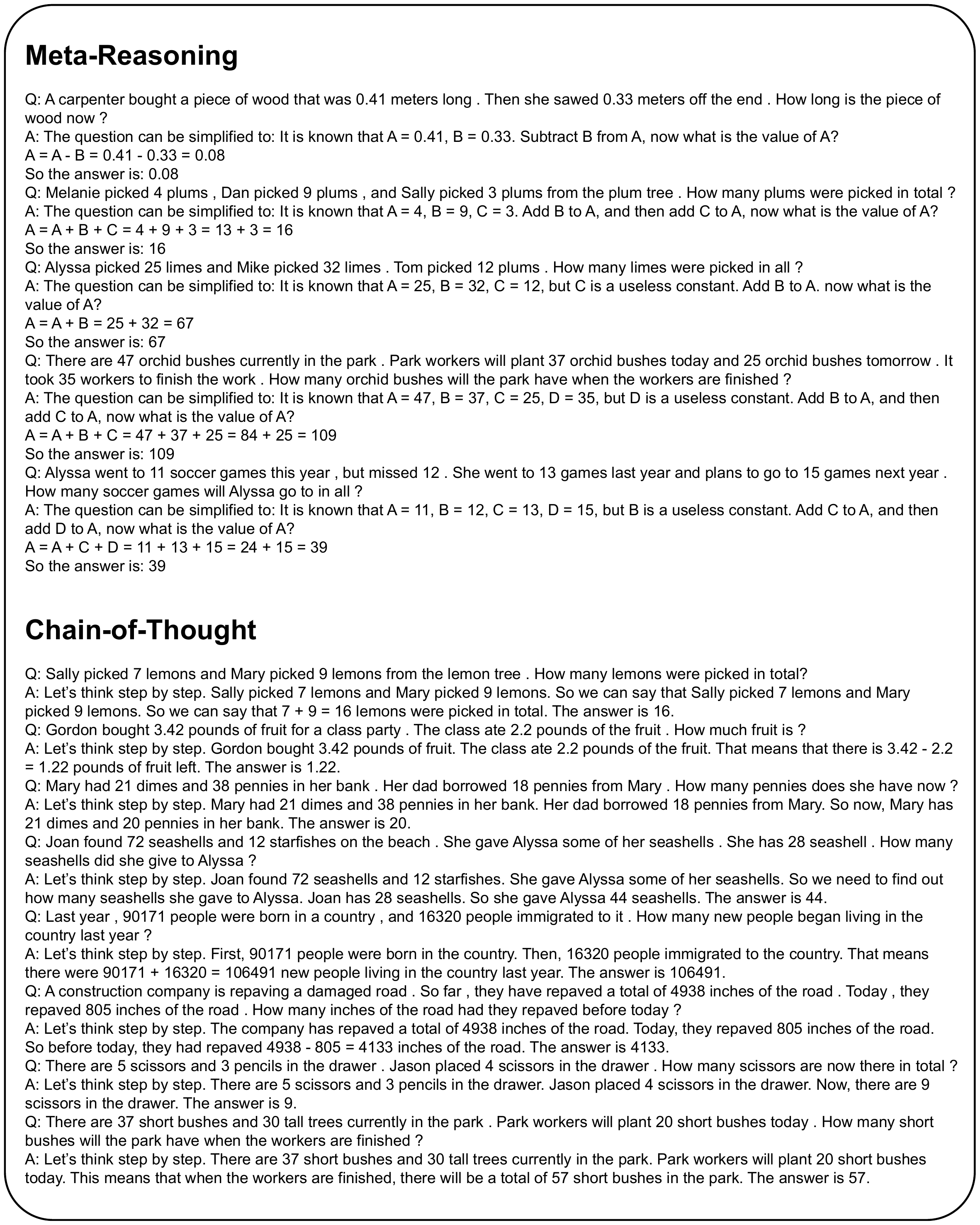}  
  \vspace{-0.25in}
  \caption{Demos: AddSub.}
  \label{Demos: AddSub}
\end{figure*}

\begin{figure*}
  \centering
  \includegraphics[width=\textwidth]{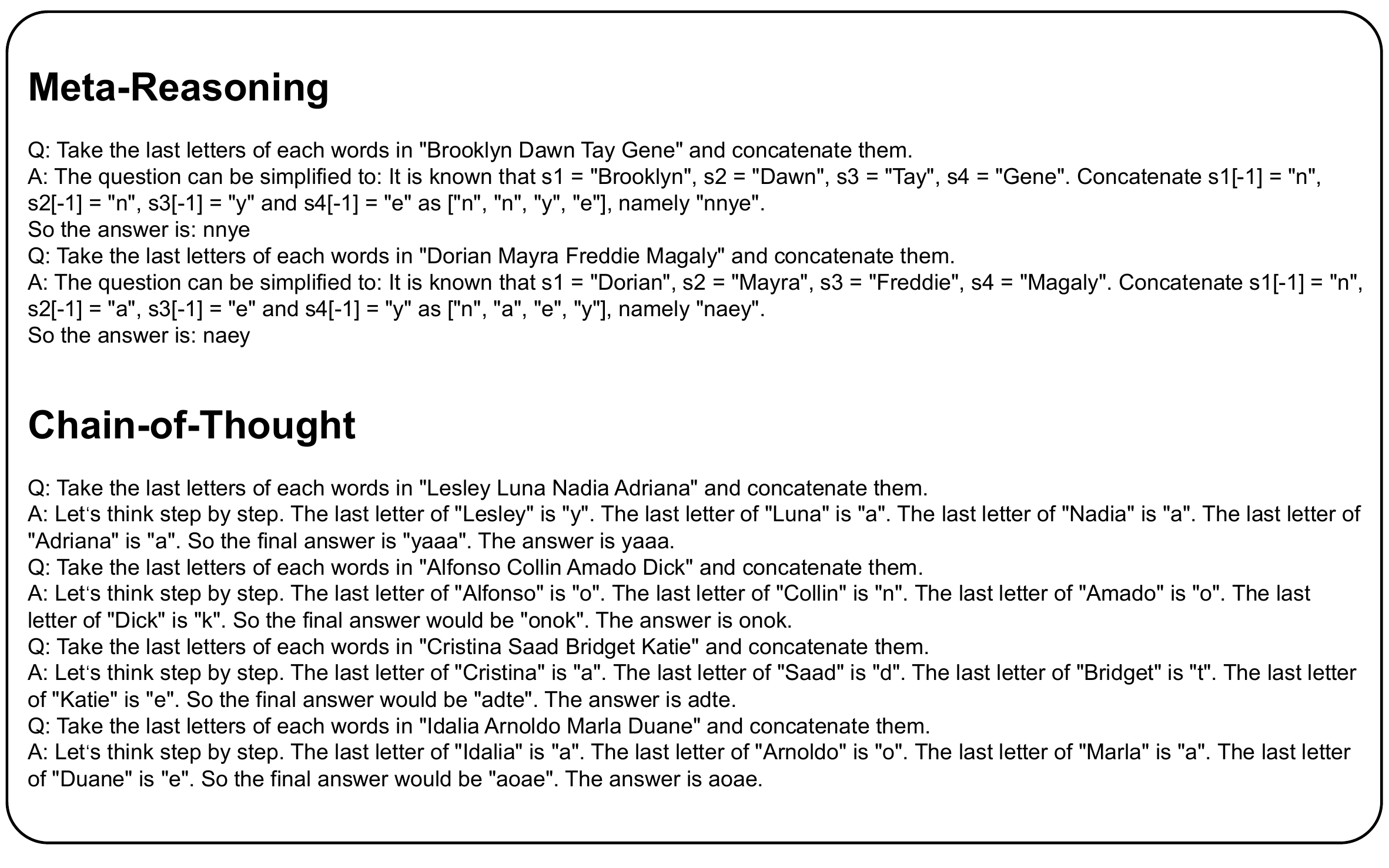}  
  \vspace{-0.25in}
  \caption{Demos: Last Letter Concatenation.}
  \label{Demos: LLC}
\end{figure*}

\begin{figure*}
  \centering
  \includegraphics[width=\textwidth]{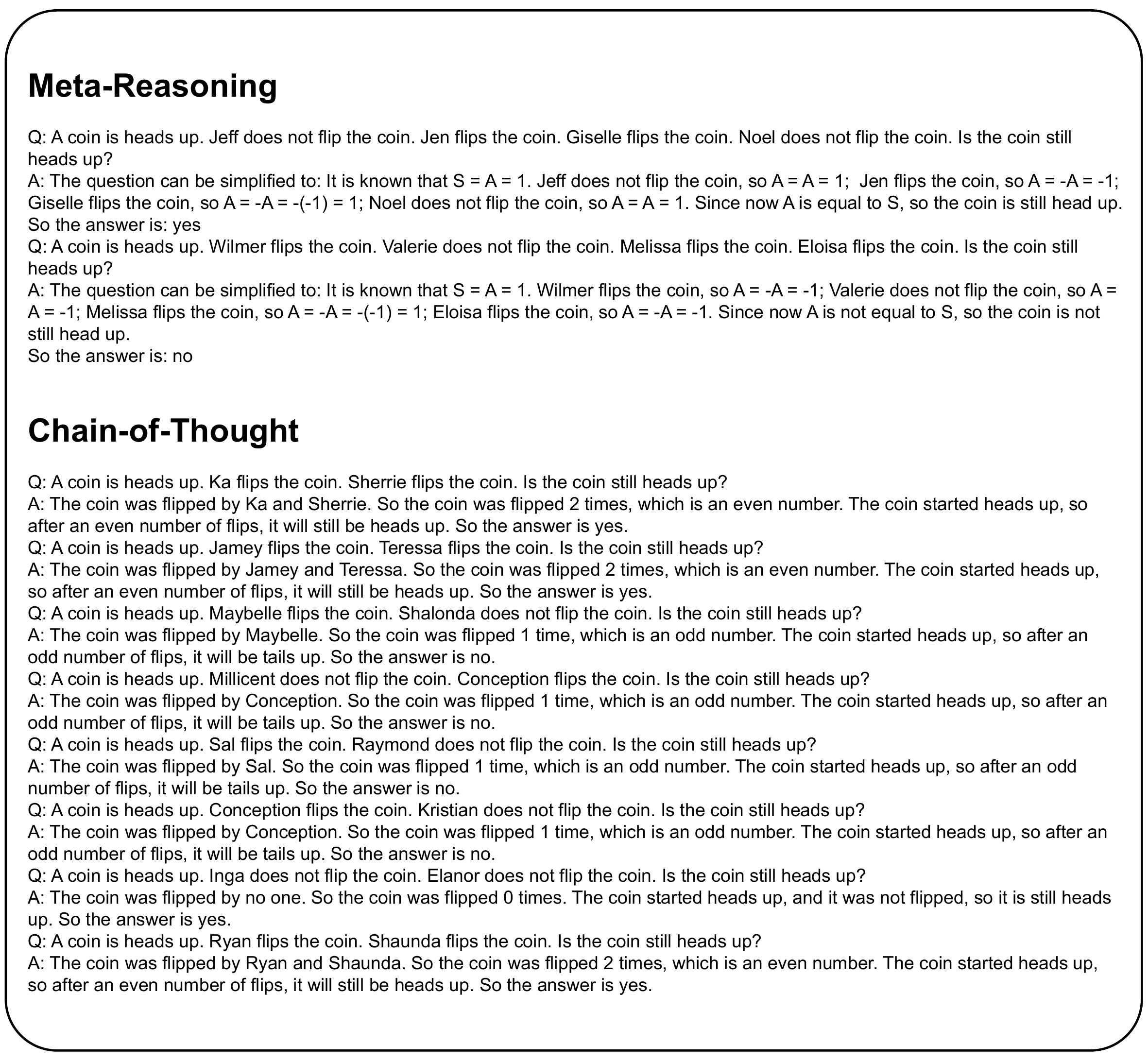}  
  \vspace{-0.25in}
  \caption{Demos: Coin Flip.}
  \label{Demos: coin}
\end{figure*}

\begin{figure*}
  \centering
  \includegraphics[width=\textwidth]{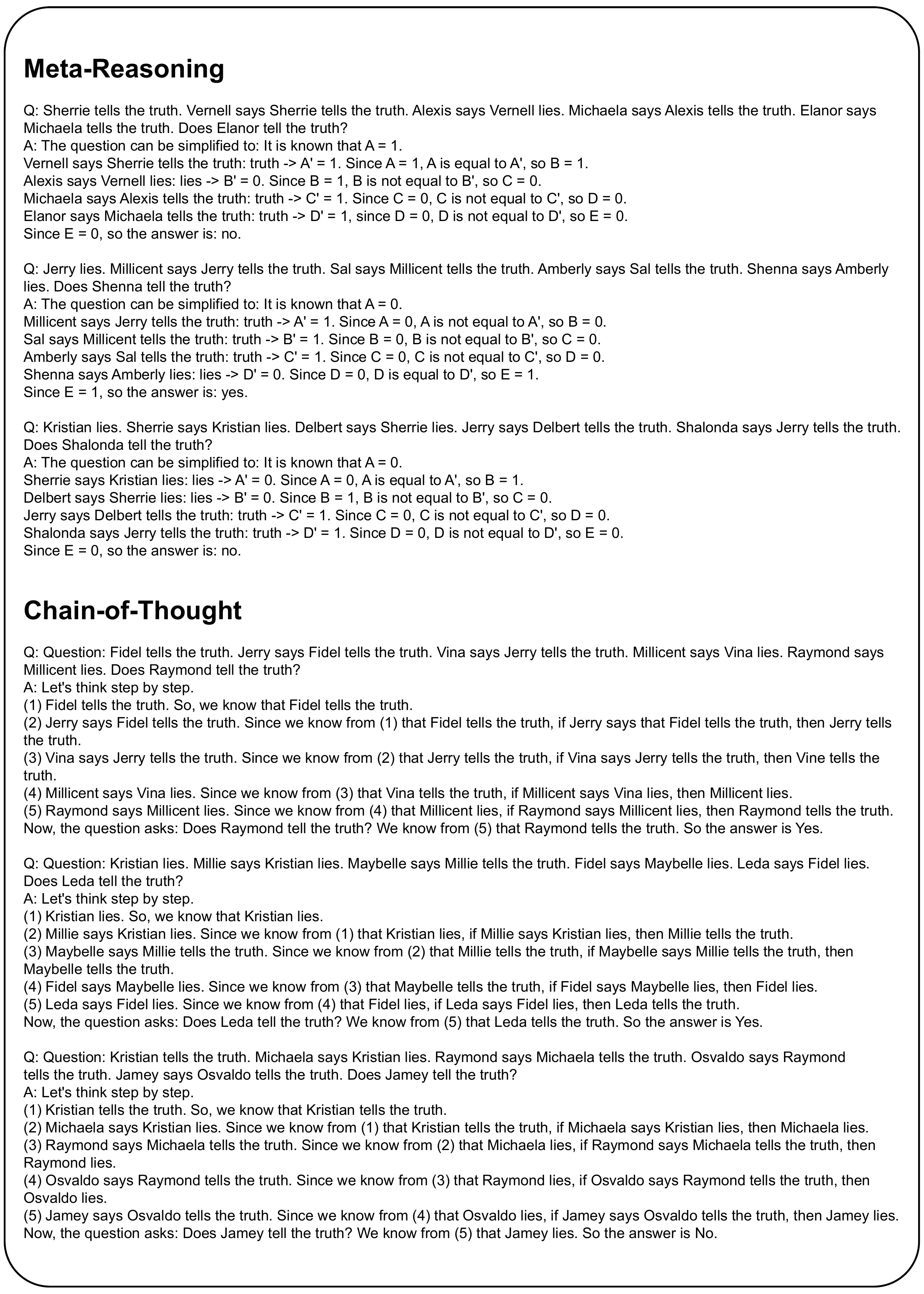}  
  \vspace{-0.25in}
  \caption{Demos: Web of Lies.}
  \label{Demos: Lies}
\end{figure*}

\begin{figure*}
  \centering
  \includegraphics[width=\textwidth]{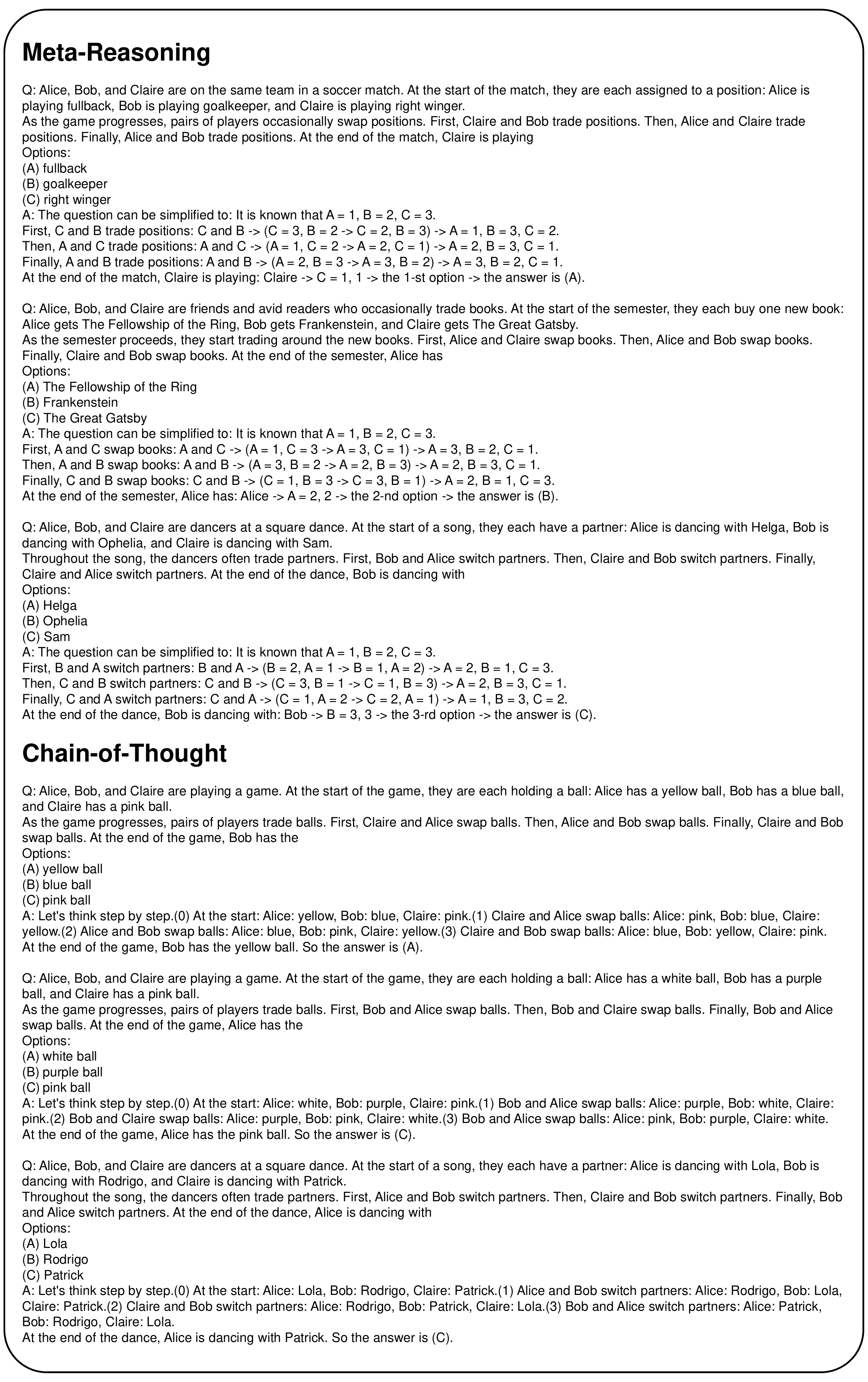}  
  \vspace{-0.25in}
  \caption{Demos: Tracking Shuffled Objects.}
  \label{Demos: Track}
\end{figure*}

\begin{figure*}
  \centering
  \includegraphics[width=\textwidth]{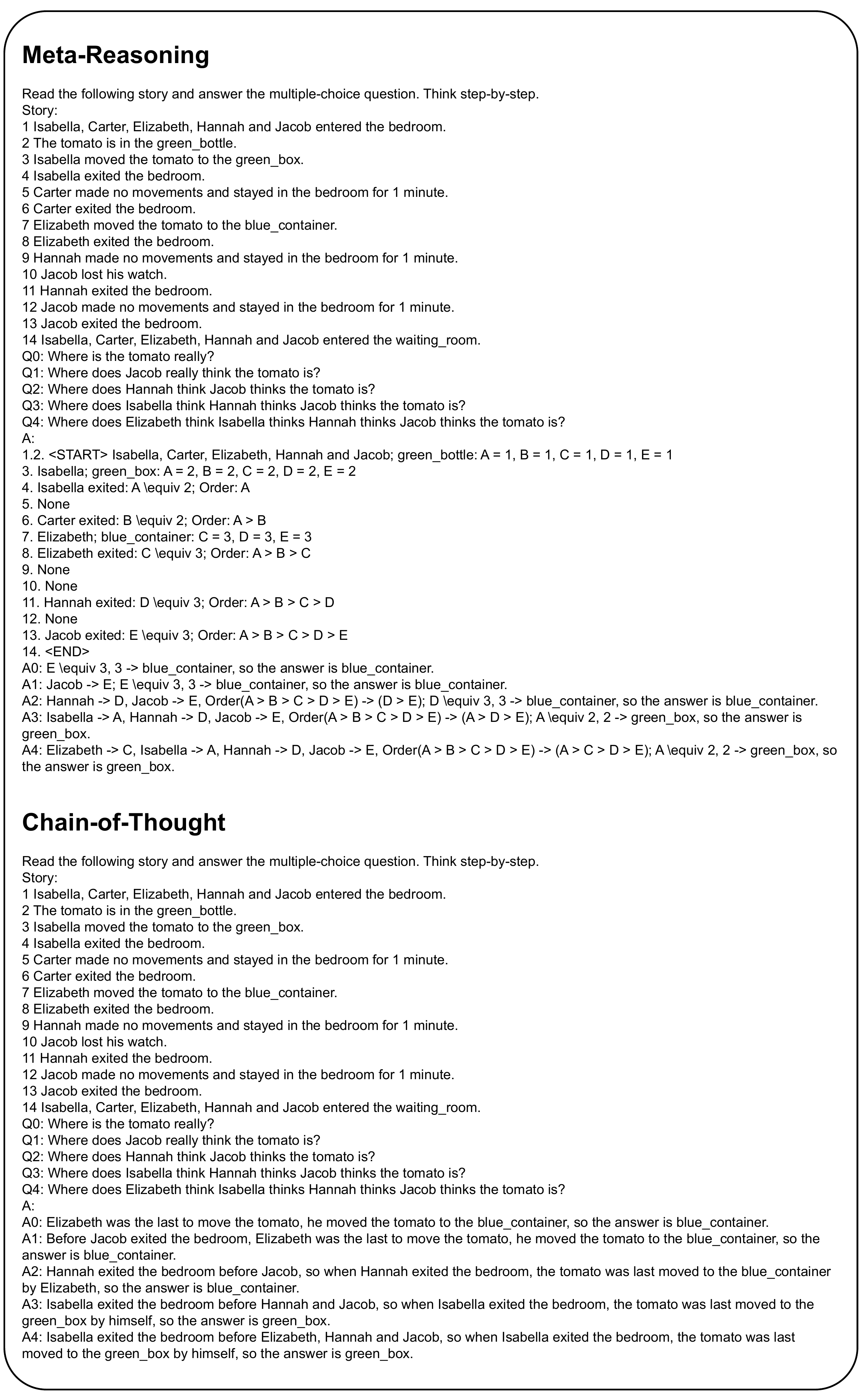}  
  \vspace{-0.25in}
  \caption{Demos: Theory-of-Mind.}
  \label{Demos: ToM}
\end{figure*}

Figure \ref{Demos: MultiArith} to \ref{Demos: ToM} show all the demonstrations used in the dataset of this paper.

\end{document}